\documentclass[10pt, journal]{IEEEtran}
\IEEEoverridecommandlockouts
\usepackage{cite}
\usepackage{amsmath,amssymb,amsfonts}
\usepackage{algorithmic}
\usepackage{graphicx}
\usepackage{textcomp}
\usepackage{xcolor}
\def\BibTeX{{\rm B\kern-.05em{\sc i\kern-.025em b}\kern-.08em
    T\kern-.1667em\lower.7ex\hbox{E}\kern-.125emX}}

\usepackage{subcaption}
\usepackage{multirow}
\usepackage{array}
\usepackage[linesnumbered,lined,commentsnumbered, ruled]{algorithm2e}

\usepackage{hyperref} 
\hypersetup{
            colorlinks,
            linkcolor=black,
            anchorcolor=black,
            citecolor=black,
            urlcolor=black}
            
\usepackage{tikz} 
\newcommand*{\circled}[1]{\lower.7ex\hbox{\tikz\draw (0pt, 0pt)%
    circle (.45em) node {\makebox[1em][c]{\footnotesize #1}};}}
    
\begin{document}

\title{\huge Tight Compression: Compressing CNN Through Fine-Grained Pruning and Weight Permutation for Efficient Implementation
\vspace{2pt}
}

\author{Xizi Chen, 
Jingyang Zhu, 
Jingbo Jiang, 
and Chi-Ying Tsui
\vspace{-20pt}

\thanks{Xizi Chen, Jingbo Jiang, and Chi-Ying Tsui are with ACCESS - AI Chip Center for Emerging Smart Systems, Hong Kong Science Park, Hong Kong, China, and the Department of Electronic and Computer Engineering, Hong Kong University of Science and Technology (HKUST), Hong Kong, China.
E-mail: xchenbn@connect.ust.hk, jjiangan@connect.ust.hk, eetsui@ust.hk
}%
\thanks{Jingyang Zhu is with NVIDIA Corporation, Shanghai, China.
E-mail: jingyangz@nvidia.com}
\thanks{This work has been submitted to the IEEE for possible publication. Copyright may be transferred without notice, after which this version may no longer be accessible.}
}

\maketitle

\begin{abstract}
The unstructured sparsity after pruning poses a challenge to the efficient implementation of deep learning models in existing regular architectures like systolic arrays. On the other hand, coarse-grained structured pruning is suitable for implementation in regular architectures but tends to have higher accuracy loss than unstructured pruning when the pruned models are of the same size.
In this work, we propose a model compression method based on a novel weight permutation scheme to fully exploit the fine-grained weight sparsity in the hardware design.
Through permutation, the optimal arrangement of the weight matrix is obtained, and the sparse weight matrix is further compressed to a small and dense format to make full use of the hardware resources.
Two pruning granularities are explored. In addition to the unstructured weight pruning, we also propose a more fine-grained subword-level pruning to further improve the compression performance.
Compared to the state-of-the-art works, the matrix compression rate is significantly improved from 5.88$\times$ to 14.13$\times$. As a result, the throughput and energy efficiency are improved by 2.75 and 1.86 times, respectively.
\end{abstract}


\bstctlcite{IEEEexample:BSTcontrol} 

\vspace{-7pt}
\section{Introduction}
\vspace{0pt}
Convolutional Neural Networks (CNNs) have achieved impressive progress over the past few years in various domains, such as image classification\cite{Alexnet}, object detection\cite{Yolo}, and semantic segmentation\cite{Unet}. At the same time, the neural network models are becoming deeper and heavier to achieve better performance. The huge number of weights increases the requirement in hardware resources and the energy consumption of inference, thus making it challenging for implementing CNNs on embedded systems. In recent works, pruning techniques have been studied to address this limitation \cite{HanCompression,ToPruneOrNot}. It is observed that contemporary neural networks tend to be over-parameterized, and a large portion of weights can be removed to reduce the model complexity and the heavy cost of inference. This pruning process only has a marginal impact on accuracy after fine-tuning the remaining weights. For example, 89\% of weights were pruned in the AlexNet \cite{Alexnet} model without accuracy loss in \cite{HanCompression}. 

Although great progress has been made in network pruning, such a high pruning rate does not directly lead to the same degree of energy saving and throughput improvement in the hardware accelerators. Many state-of-the-art CNN accelerators employ systolic arrays to implement the intensive multiply-accumulate computations (MACs) during inference \cite{TpuGoogle,PackCNN}. As one of the most popular architectures to implement CNN models, the systolic array has the advantage of regular structure, parallel computation, and data reuse capability. Therefore, it can execute the intrinsic matrix multiplications of CNN very efficiently with high energy efficiency and throughput. However, it is not easy for the regular structure of the systolic array to take full advantage of the fine-grained sparsity in the pruned network models. Since the nonzero weights have an irregular distribution in the weight matrix, the size of the weight matrix of each layer cannot be reduced efficiently by the unstructured pruning. Therefore, most of the zero weights will still be mapped to the systolic array nodes, making it difficult to improve the throughput and energy efficiency. As one potential way to tackle this problem, structured pruning techniques have been explored \cite{NetworkSlimming,FilterPruning1,FilterPruning2}, where the pruning is performed in a larger granularity like channel-wise and kernel-wise. However, structured pruning usually tends to have higher accuracy loss than the fine-grained unstructured pruning when the pruned models are of the same size \cite{TheStateOfSparsity,PackCNN}.

In this work, we propose a tight compression method to compress the unstructurally-pruned CNN model to a small and dense format, so that the systolic array can be fully utilized in implementing the pruned neural network. Compared to the state-of-the-art model compression techniques, the proposed method can achieve a higher compression rate of the weight matrix, which leads to significant improvements in throughput and energy efficiency. In summary, this work makes the following contributions: 
\begin{itemize}\setlength{\itemsep}{2pt}
\item A compression method is proposed to implement CNN models efficiently in the systolic array system. The unstructured pruning is carried out first. Then, the sparse weight matrix is partitioned according to the size of the systolic array, and the rows and columns are permuted to facilitate packing. Simulated annealing (SA) is used during the permutation process to obtain the optimal arrangement of the weight matrix for the final compression step. After permutation, the sparse weight matrix can be compressed to a small and dense format to make full use of the hardware resources. 

\item Two pruning granularities are explored for weight permutation and matrix compression. The design based on the unstructured weight pruning is presented first (\textit{i.e.} weight-level compression). Then, the model is further pruned at the subword level to exploit the fine-grained subword sparsity for improving the compression performance (\textit{i.e.} subword-level compression). 

\item The hardware structure of the systolic array is taken into account during the compression process. Instead of only considering the pruning rate, we adopt the compression rate of the weight matrix as the optimization objective for the simulated annealing-based weight permutation.

\item A systolic array architecture is designed for implementing the compressed CNN models. Experimental results show that the proposed compression method can prune over 93\% of the weights and compress the size of the weight matrices by 14.13 times. As a result, significant improvements can be achieved in throughput and energy efficiency.
\end{itemize}

This work is an extension of our prior paper \cite{TightCompression}.
The extended materials include 1) a more fine-grained subword-level compression method to further improve the compression performance, and 2) the hardware architecture to support the subword-level compressed models.

\section{Preliminaries}
\label{sec: preliminaries}

\begin{figure}[!t]
\setlength{\belowcaptionskip}{-4pt}
\centering
	\includegraphics[width=0.482\textwidth]{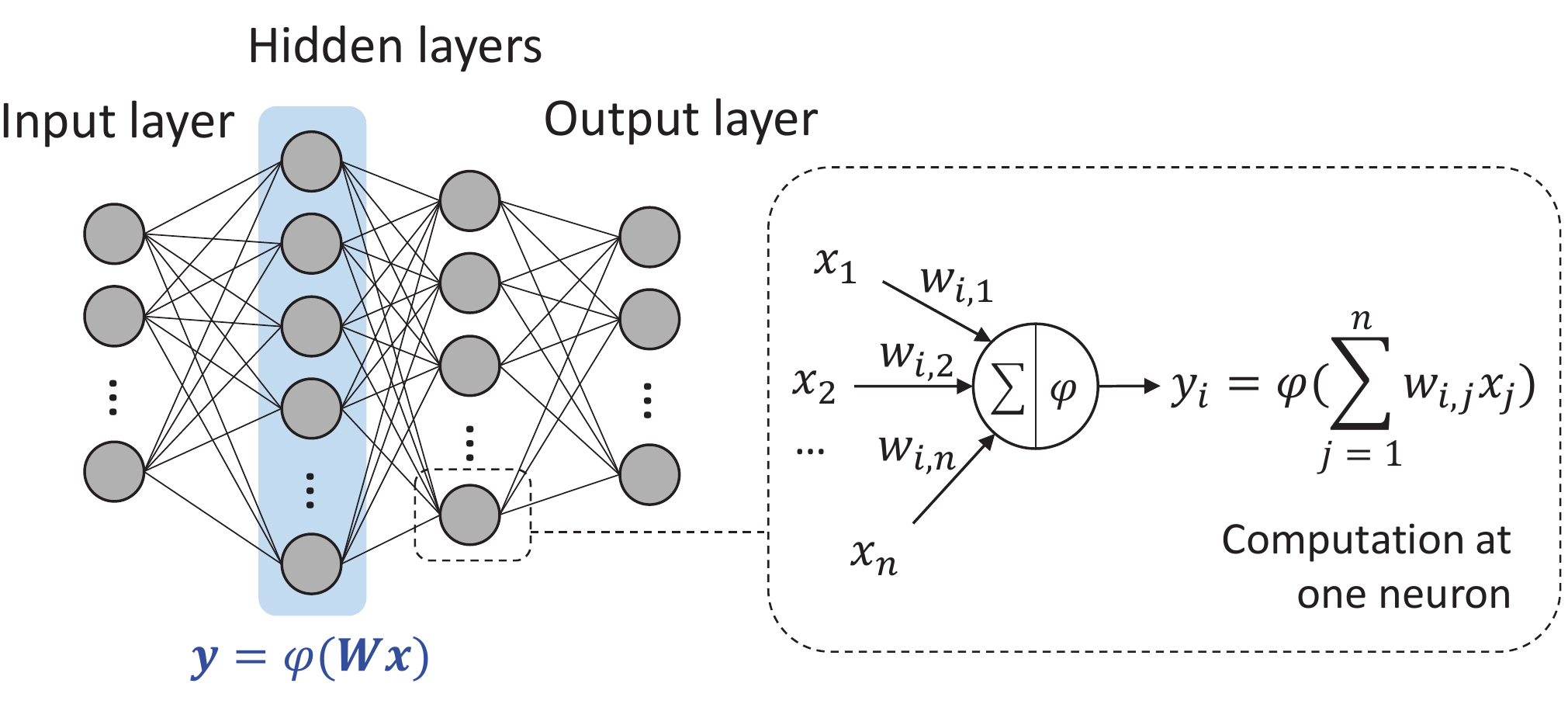} 	
	\caption{CNN Model and the MAC Operation at One Neuron.}
	\label{fig: cnn}
\end{figure}

\subsection{CNN Model and Systolic Array} 
\label{subsec: systolic_array}
Convolutional neural network (CNN) is a machine learning model inspired by the brain. It is organized as a stack of layers to extract the features hierarchically, as shown in Fig.~\ref{fig: cnn}. Convolutional (CONV) layers are the core layers for feature processing and thus account for most of the arithmetic computations. At each neuron, the weighted sum of the input activations is computed. Then, the multiplication and accumulation (MAC) result is sent to the activation function to calculate the output activation. Since the intrinsic computation at each layer can be regarded as a matrix multiplication between weights and input activations, CNN models are highly suitable to be implemented using systolic arrays. 

Fig.~\ref{fig: systolic_mult}~(a) shows the general architecture of a systolic array. Each node in the systolic array is a processing unit connected with the four neighboring nodes. In the widely used weight-stationary computation flow, the weight matrix is stored in the local registers of the nodes and will not move during the matrix multiplication \cite{TpuGoogle,PackCNN}. Each row in the systolic array is mapped with the weights of the same neuron. The input activations are sent into the array from the bottom and gradually propagate to the top. As the input activations propagate across the nodes, the weight stored in each node is continually multiplied with the input activations received from the bottom node. The product is accumulated with the corresponding partial sum received from the left node and then propagates to the right. All the weight-activation products related to the same output will be summed up along the row and sent out of the systolic array from the right. 

\subsection{Related Works}
\label{subsec: related_works}

\subsubsection{Pruning}
Network pruning seeks to minimize the number of nonzero weights in the over-parameterized CNN model to reduce the model complexity and the cost of inference. In \cite{HanCompression}, Han \textit{et al.} propose to prune the weights according to their magnitude. Weights with small magnitude values are considered to have a relatively small contribution to the model quality and thus can be removed. After pruning, the remaining weights are retrained to recover the accuracy. The method can achieve a pruning rate of 89\% in AlexNet \cite{Alexnet}. In \cite{ToPruneOrNot}, an automated pruning algorithm is proposed to gradually prune the small-magnitude weights to a preset level of sparsity with minimal retraining requirement. Considerable pruning rates have been achieved in previous works.
However, it is difficult to fully exploit the unstructured sparsity in the regular hardware architectures like systolic arrays. Since the zero weights are randomly spread across the matrix, the size of the weight matrix cannot be reduced efficiently by unstructured pruning, and thus most zero weights still have to be allocated to the nodes of the systolic array. The nodes mapped with zero weights will not perform effective computations, thus affecting the overall improvements in throughput and energy efficiency. 

\subsubsection{Structured Pruning}
Structured pruning techniques have been explored in recent works to implement the sparse neural networks in the existing regular architectures \cite{NetworkSlimming,FilterPruning1,FilterPruning2}. Instead of pruning individual weights, the structured pruning is performed in a larger granularity such as an entire row or column in the weight matrix. After pruning, the model will have a structured sparsity that can be mapped to the systolic array efficiently. However, the number of nonzero weights after the structured pruning is usually larger than the fine-grained pruning if the same level of accuracy is maintained \cite{TheStateOfSparsity,PackCNN}. For example, 4.1M nonzero weights are preserved after the unstructured pruning \cite{PackCNN} on the ImageNet dataset \cite{ImageNet}, whereas 23.2M nonzero weights are kept after the structured pruning \cite{NetworkSlimming}. 

\begin{figure}[!t]
\setlength{\belowcaptionskip}{-8pt}
\centering
	\includegraphics[width=0.485\textwidth]{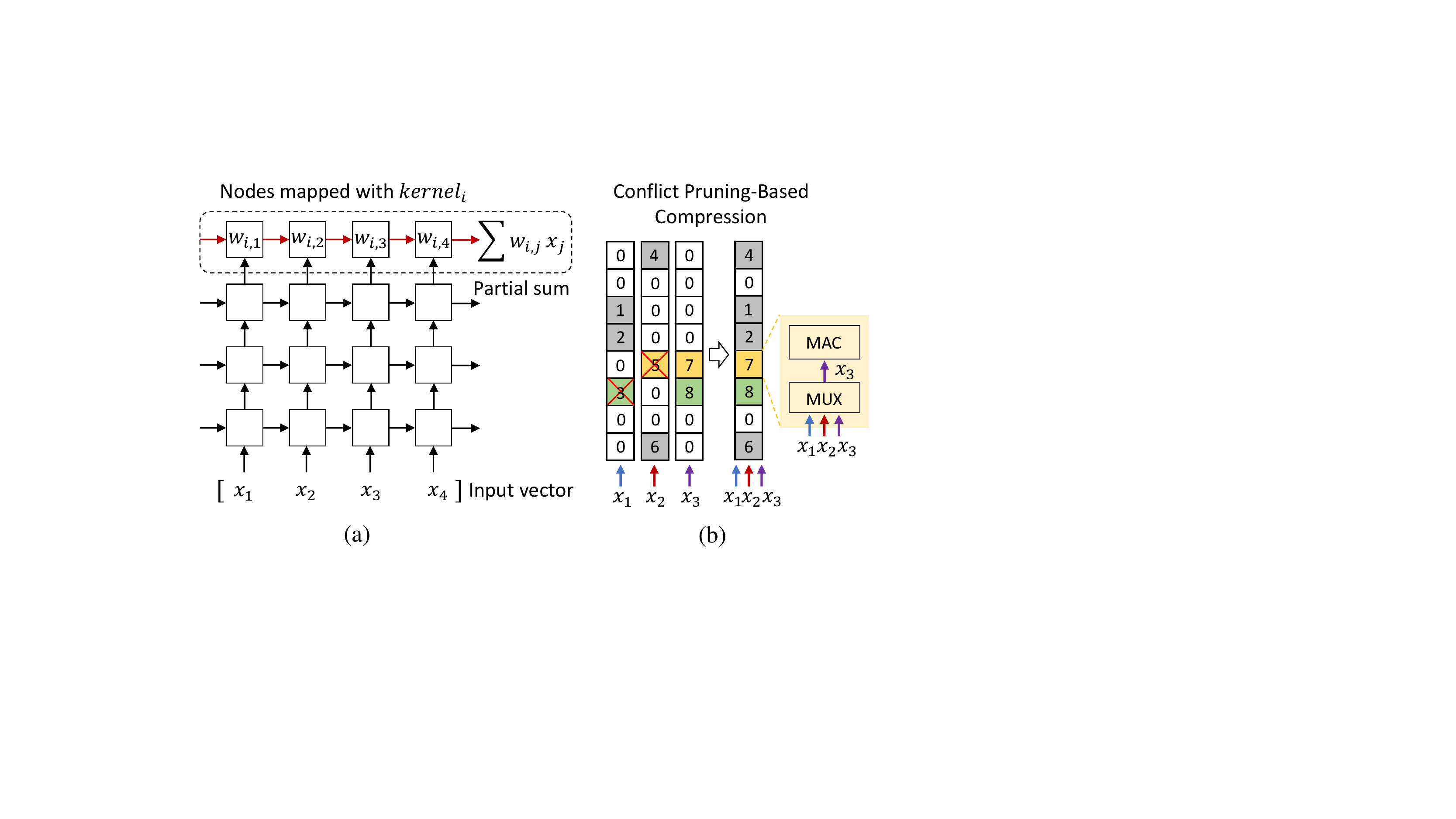} 	
	\caption{(a) Matrix Multiplication in the Systolic Array and (b) the Conflict Pruning-Based Compression \cite{PackCNN}.}
	\label{fig: systolic_mult}
\end{figure}

\subsubsection{Column Packing and Conflict Pruning}
Another way to map the sparse networks efficiently to the systolic array is to reduce the size of the sparse weight matrix by packing the columns. A novel method is proposed in \cite{PackCNN}. The moderate unstructured pruning is carried out first. Then, different weight columns are grouped, so that later the entire group can be mapped to a single column of the systolic array to save the energy and execution time. Inside the group, only one nonzero weight is allowed at each row position to make sure that each node only needs to store one weight and handle one MAC operation at a time. To perform matrix multiplication, the input activations also need to be grouped and sent to the corresponding columns of the systolic array. At each node, only one activation will be selected to multiply with the weight. Since the partial results are accumulated along the rows, column packing will not cause any error or hardware overhead to the accumulation process. However, compared to the conventional unstructured pruning, extra accuracy loss will be induced by a second-time pruning called \textbf{conflict pruning}. Since there are usually hundreds and thousands of rows in the weight matrix, most columns very likely have some nonzeros at the same row positions. These nonzeros are called conflicts between the columns. Directly packing the weight matrix only has a limited benefit in matrix compression as it is difficult to find a set of columns that has no conflict. To make the packing more efficient, a specific number of conflicts are allowed in each column group in \cite{PackCNN}. The model will then undergo a second pruning step, where all the conflicts except the one with the largest magnitude at each row are pruned in every column group. This compression method is illustrated in Fig.~\ref{fig: systolic_mult}~(b). Two pairs of conflicts exist in the column group, and the nonzero weights 3 and 5 will be pruned during conflict pruning. After mapping the column group to the systolic array, the appropriate input will be selected at each node to compute with the weight, as shown in Fig.~\ref{fig: systolic_mult}~(b).

\subsubsection{Mixed-Precision Computation}
In addition to the pruning techniques, mixed-precision computation is another way to reduce the computational complexity \cite{BiScale, OutlierAware}. Since most weights have small values, and only a small portion of weights are orders of magnitude larger, a smaller bit-length (\textit{e.g.} $<$8-bit) can be achieved by using mixed-precision to represent the weights.
In \cite{BiScale}, two different scale factors are used for quantizing the weights at a layer. For weights with small magnitude, a narrow quantization range with high resolution is used. On the contrary, for weights with large magnitude, a wide quantization range with low resolution is adopted. A smaller bit-length (\textit{e.g.} 6-bit) is used for all weights.
In \cite{OutlierAware}, the large weights are called outliers and represented using a large bit-length (\textit{e.g.} 16-bit). The majority of weights with small magnitude adopt a 4-bit representation to reduce the computations.
This mixed-precision computation usually requires special hardware design for implementation. For example, different from the normal computations that are handled by 4-bit MAC units, the outliers require dedicated high-precision processing elements to compute in \cite{OutlierAware}.

\section{Weight-Level Tight Compression and Accelerator Design}
\label{sec: tight_compression}

\subsection{Overview}
To make the column packing efficient, the number of conflicts allowed for each group is equal to $1.75R$ in \cite{PackCNN}, where $R$ is the number of rows in the weight matrix. When $R$ is equal to 1k, 1750 conflicts are allowed per group. Such a large amount of conflict pruning has a non-trivial impact on the model quality. As a result, more nonzero weights have to be preserved to maintain high accuracy, compared to the efficient unstructured pruning. For example, 0.3M nonzero weights are preserved after conflict pruning with an accuracy of 92.9\% on the CIFAR-10 dataset \cite{Cifar10}. However, the efficient unstructured pruning can reduce the number of nonzero weights to 0.13M in the same neural network with an accuracy of 93.10\%.

Since the weight matrix usually has a much larger size than the systolic array, it needs to be partitioned into smaller weight tiles and mapped to the systolic array multiple times to finish the entire matrix multiplication.
In this work, we leverage this weight tiling and propose a novel weight permutation method to avoid conflicts in the potential groups and facilitate column packing without sacrificing accuracy.
Different from \cite{PackCNN}, no conflict pruning is performed, and thus a more aggressive pruning rate is achieved with high accuracy. Through permutation, the size of the weight matrix of each layer can be significantly reduced compared to conflict pruning. As a result, fewer weight tiles need to be mapped to the systolic array for computation, and thus higher throughput and energy efficiency can be achieved.

\begin{algorithm}[tb]
\small
\SetKwFunction{PruneSchedule}{prune-schedule}
\SetKwFunction{MagnitudePrune}{magnitude-prune}
\SetKwFunction{train}{train}
\SetKwFunction{MatrixExtract}{matrix-extract}
\SetKwFunction{SaOptimize}{matrix-compress}
\KwIn{Network model ($M$); Epochs ($E$); Pruning rate ($P$); \\
\hskip3.0em Max. number of columns per group ($G$)}
\KwOut{Compressed network model ($M_c$)}
\textbf{Stage 1:  Unstructured Pruning} \\
$e,~p \leftarrow$ \PruneSchedule{$E,P$} \;
$M_c \leftarrow M$ \;
\For{$i \leftarrow 1$ \KwTo $E$} {
	\If{$i \in e$} {
		$M_c \leftarrow$ \MagnitudePrune{$M_c,p_i$} \;
	}
	$M_c \leftarrow$ \train{$M_c$} \;
}
\textbf{Stage 2:  Weight Matrix Compression}\\
\For{$ l \leftarrow 1$ \KwTo $layers$} {
	$m_l \leftarrow$ \MatrixExtract{$M_c,l$} \;
	$m_l \leftarrow$ \SaOptimize{$m_l,G$} \;
}
\small
\caption{Tight Compression Overview}
\label{alg_overview}
\end{algorithm}

The high-level flow of the proposed compression method is summarized in Algorithm~\ref{alg_overview}. In the first stage, the network model is gradually pruned up to a preset level of pruning rate ($P$) in the $E$ training epochs. Before training, the subroutine $prune$-$schedule$ is invoked in Line 2 to schedule the intermediate pruning epochs ($e$) and the corresponding pruning rates ($p$). At each pruning epoch $e_i$, the subroutine $magnitude$-$prune$ is invoked to prune the small-magnitude weights to a percentage of $p_i$ in each layer. To make an apple-to-apple comparison with the conflict pruning-based compression \cite{PackCNN}, we adopt the same pruning schedule proposed in \cite{ToPruneOrNot}. The pruning rate gradually increases in the first half of the training process. After that, the neural network is trained without further pruning to recover accuracy. After training, the compression enters the second stage, where the sparse weight matrix  ($m_l$) of each layer is further compressed to a small and dense format through weight permutation ($matrix$-$compress$). To obtain the optimal permutation result, we adopt simulated annealing algorithm for optimization. Details will be discussed shortly in the following subsections.

\subsection{Weight Permutation and Matrix Compression}
\label{subsec: weight_permutation}
\begin{figure}[!t]
\setlength{\belowcaptionskip}{-4pt}
\centering
	\includegraphics[width=0.485\textwidth]{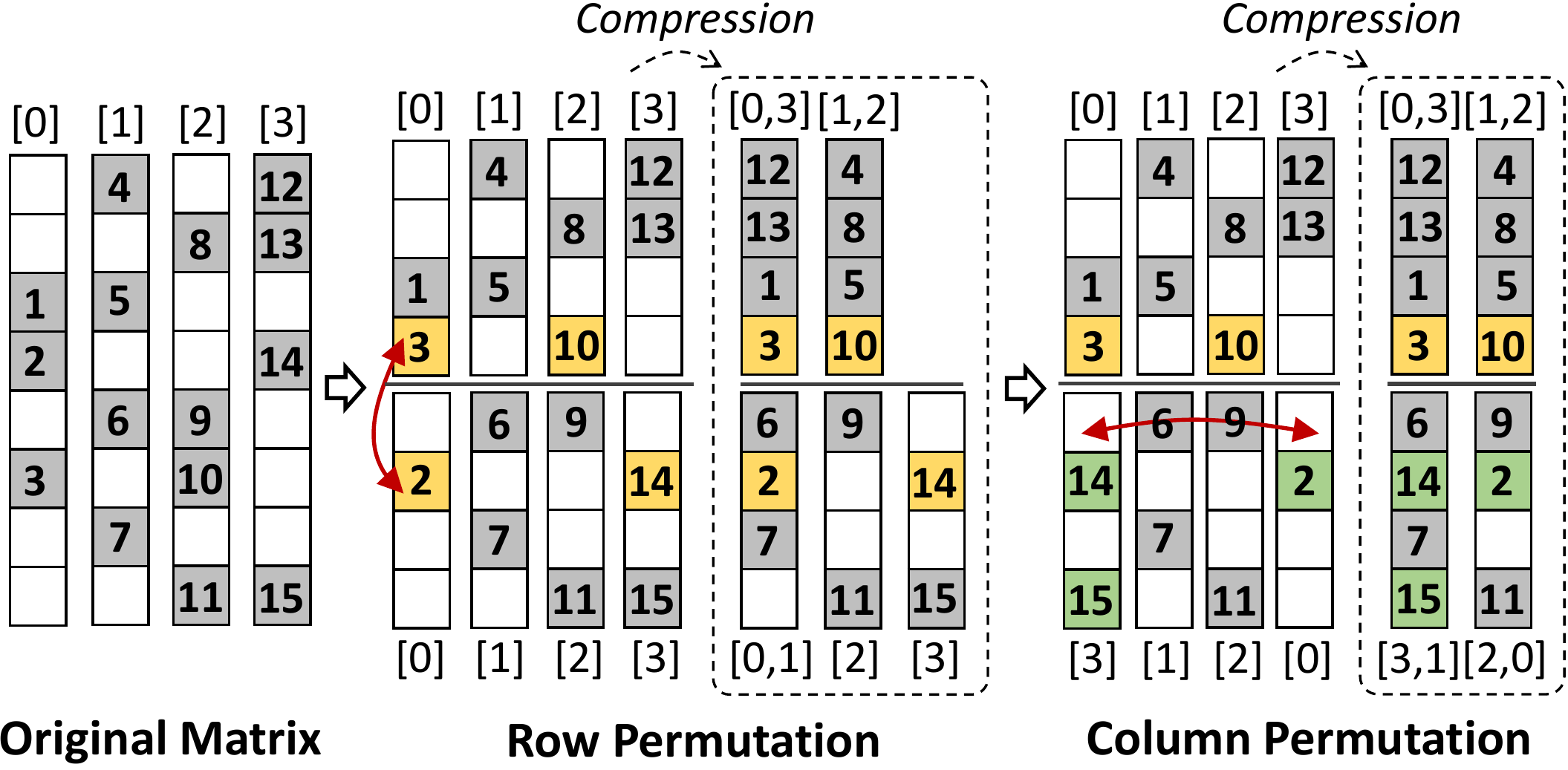} 	
	\caption{An Illustrating Example of Weight Permutation.}
	\label{fig: perm_toy_example}
\end{figure}

As mentioned in Section \ref{subsec: related_works}, directly grouping the columns without conflict pruning cannot compress the weight matrix efficiently, since any conflicts among the columns will make the packing invalid. The intuition of weight permutation is to compress the weight matrix by partitioning it into several sub-matrix sections according to the size of the systolic array and permuting the rows and columns across different sub-matrices to avoid conflicts as much as possible for efficient column packing in each sub-matrix section. As the weight matrix is usually much larger than the systolic array, we need to divide it into tiles and map each tile onto the array for computation. The weights of a column group of a tile are loaded into the corresponding column of the array. Therefore, we only need to make sure there is no conflict among the group columns of each tile instead of the group columns of the whole original weight matrix. It gives much higher flexibility to carry out the matrix compression.
An illustrating example is shown in Fig.~\ref{fig: perm_toy_example}. The weight matrix originally has a size of $8\times4$, and we assume the size of the target systolic array is $4\times4$. For clarity, only the nonzero weights are shown in the figure. Since each pair of the columns has at least one conflict, the columns cannot be grouped directly. For weight permutation, the matrix is firstly divided into \textbf{row sections}. The height of each row section is equal to the number of rows in the systolic array, so that different row sections will not be mapped to the systolic array at the same time. In this case, the columns in each section can be grouped independently to maximize the overall compression rate. 

To explain how permutation works, a one-step \textbf{row swapping} is performed between the two row sections, as shown in Fig.~\ref{fig: perm_toy_example}. Then the permuted matrix is compressed using Algorithm~\ref{alg_col_packing}. In the beginning, each column group in each row section ($m_{sec}$) only has one column entry. The packing starts from the first group $g$ ($index_g=0$) and searches the other columns to find the one ($g'$) that has no conflicts with $g$ and, at the same time, can achieve the densest format (\textit{i.e.} with the minimum number of zeros) after merging with $g$ (Line 7). If $g'$ exists, the two groups will be merged into a new group $g$ in Line 9. In the case that multiple candidates with the same density exist, the first one will be combined with $g$. This process will repeat until the current $g$ can no longer combine with any columns since there are conflicts or the number of columns in $g$ reaches the upper bound ($G$) that can be supported by the systolic array. Then the next column group will be processed (Line 11). After compression, the number of zero weights is reduced from 17 to 5 in Fig.~\ref{fig: perm_toy_example}, and the total size of the weight matrix is reduced by 37.5\% (from $8\times4$ to $4\times2$ and $4\times3$). 
The second row section can still be further compressed. Since the packing starts from the left, a different column order will lead to different compression results. To exploit this phenomenon, the columns in the row sections are permuted to search for the optimal order for packing. For example, a one-step \textbf{column permutation} is performed on the second row section in Fig.~\ref{fig: perm_toy_example}. After compression, the second row section only has 2 column groups. The number of zero weights is reduced from 17 to 1, and the size of the weight matrix is reduced by 50\%. 

\begin{algorithm}[tb]
\small
\SetKwFunction{SectionExtract}{section-extract}
\SetKwFunction{Width}{width}
\SetKwFunction{GroupExtract}{group-extract}
\SetKwFunction{DensestPair}{find-densest-group}
\SetKwFunction{Merge}{merge}
\KwIn{Permuted weight matrix ($m$); Max. number of columns per group ($G$)}
\KwOut{Packed weight matrix ($m_p$)}
$m_p \leftarrow m$ \;
\For{$i \leftarrow 1$ \KwTo{ $last~section$}}{
	$m_{sec} \leftarrow $ \SectionExtract{$m_p, i$} \;
	$index_{g} \leftarrow 0$ \;
	\While{$index_{g} < $ \Width{$m_{sec}$}}{
			$g \leftarrow$ \GroupExtract{$m_{sec},index_{g}$} \;
			$g' \leftarrow$ \DensestPair{$m_{sec}, g, G$} \;
			\uIf {$g' \neq \emptyset$}{
				$g \leftarrow$ \Merge{$g',g$} \;
			}			
			\Else{
				$index_{g} \leftarrow index_{g} + 1$ \;
			}
	}	
}
\small
\caption{Packing the Permuted Weight Matrix}
\label{alg_col_packing}
\end{algorithm}

\begin{figure}[!t]
\setlength{\belowcaptionskip}{-12pt}
\centering
	\includegraphics[width=0.485\textwidth]{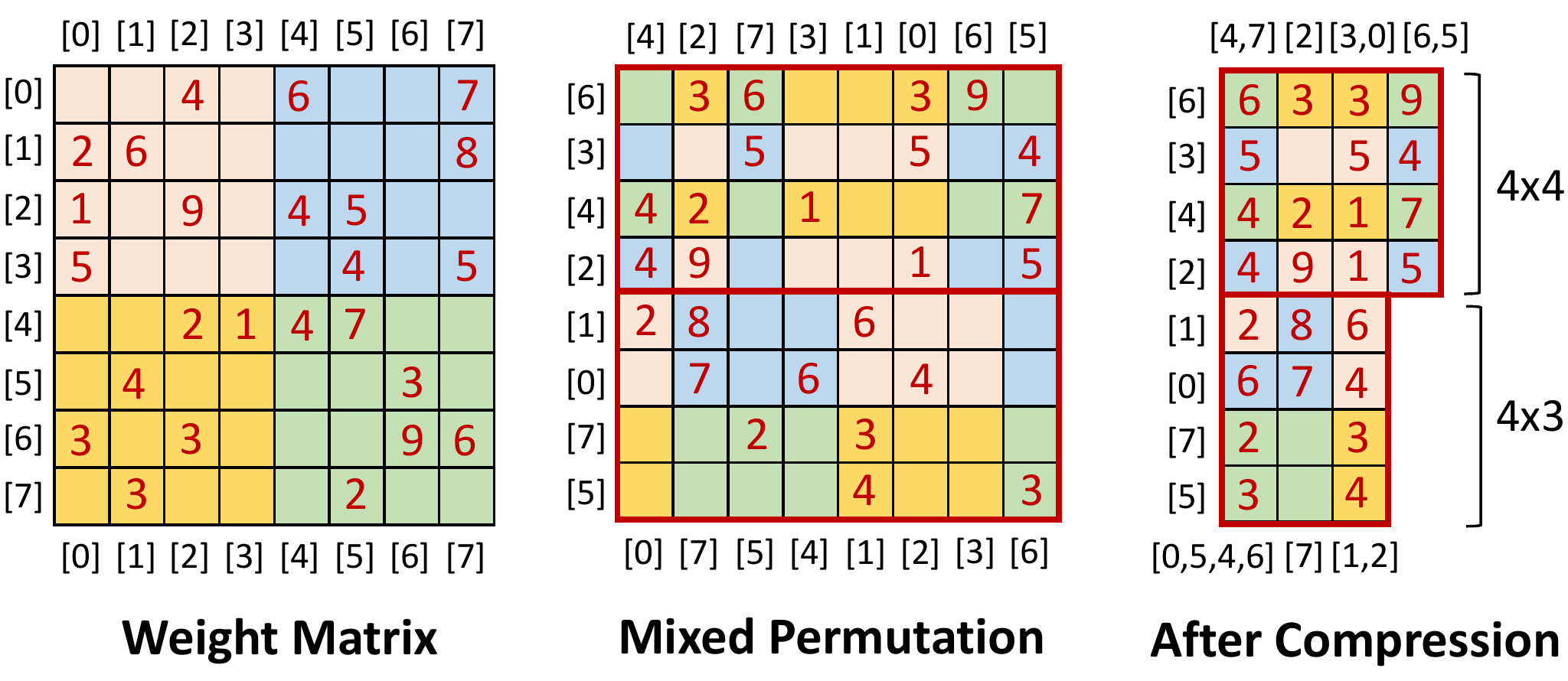} 	
	\caption{Compression Result of a Weight Matrix with 2 Row Sections.}
	\label{fig: perm_matrix}
\end{figure}

Fig.~\ref{fig: perm_matrix} shows the result after multiple steps of the mixed permutation in a small weight matrix with two row sections. The columns in each row section can be ordered independently, and each row can be permuted to either row section. Since the row permutation is across different sections, the column order of the permuted row has to follow that of the destination row section. Through permutation, the weight matrix can be compressed efficiently, reducing the number of weight tiles from 4 to 2 (each tile has a size of $4\times4$).

\subsection{Simulated Annealing Based Permutation}


Weight permutation is an effective way to improve the matrix compression rate without accuracy loss. However, finding the optimal order of rows and columns so that the compression can be optimized is computationally intractable. For example, there are $\frac{(8!)^3}{(8-4)!4!}=10^{11}$ possible states in the small matrix shown in Fig.~\ref{fig: perm_matrix}. Since the complexity increases explosively with the matrix size, it is impractical to search through the huge state space exhaustively. In this work, we adopt \textbf{simulated annealing (SA)} \cite{BinPacking,PlacementRoute} to find a close-to-optimal solution efficiently.

\begin{algorithm}[tb]
\small
\SetKwFunction{Pack}{pack}
\SetKwFunction{NeighborState}{neighbor-state}
\SetKwFunction{DeltaEnergy}{delta-energy}
\SetKwFunction{Random}{random}
\SetKwFunction{TempDecay}{temp-decay}
\KwIn{Weight matrix ($m$); Max. number of columns per group ($G$); Initial and final temperatures ($T_{init}$, $T_{end}$); Cooling factor ($f$); Number of iterations at each temperature ($iter$)}
\KwOut{Compressed weight matrix after optimization ($m_c$)}
$state\leftarrow m$ \;
$state_{p} \leftarrow$ \Pack{$state, G$} ; \tcp{\scriptsize Algorithm 2}
$temp \leftarrow T_{init}$ \;
$loops \leftarrow 0$ \;
\While{$temp > T_{end}$}{
	$state' \leftarrow$ \NeighborState{$state$};\tcp{\scriptsize{Permutation}}
	$state'_p \leftarrow$ \Pack{$state', G$} \; 
	$E'-E \leftarrow$ \DeltaEnergy{$state'_p, state_p$}	\;
	\If{\Random{$0,1$}$<e^{-\frac{E'-E}{temp}}$}	{
			$state\leftarrow state'$\;
			$state_p\leftarrow state'_p$\;
	}
	$loops\leftarrow loops+1$ \;
	\If{$loops=iter$}{
		$temp\leftarrow temp\times(1-f)$  ; \tcp{\scriptsize Cooling}
		$loops\leftarrow0$ \;
	}
}
$m_c\leftarrow state_p$ \;
\small
\caption{SA-Based Weight Permutation}
\label{alg_sa}
\end{algorithm}

\begin{figure}[!t]
\setlength{\belowcaptionskip}{-8pt}
\centering
\vspace{-6pt}
	\includegraphics[width=0.47\textwidth]{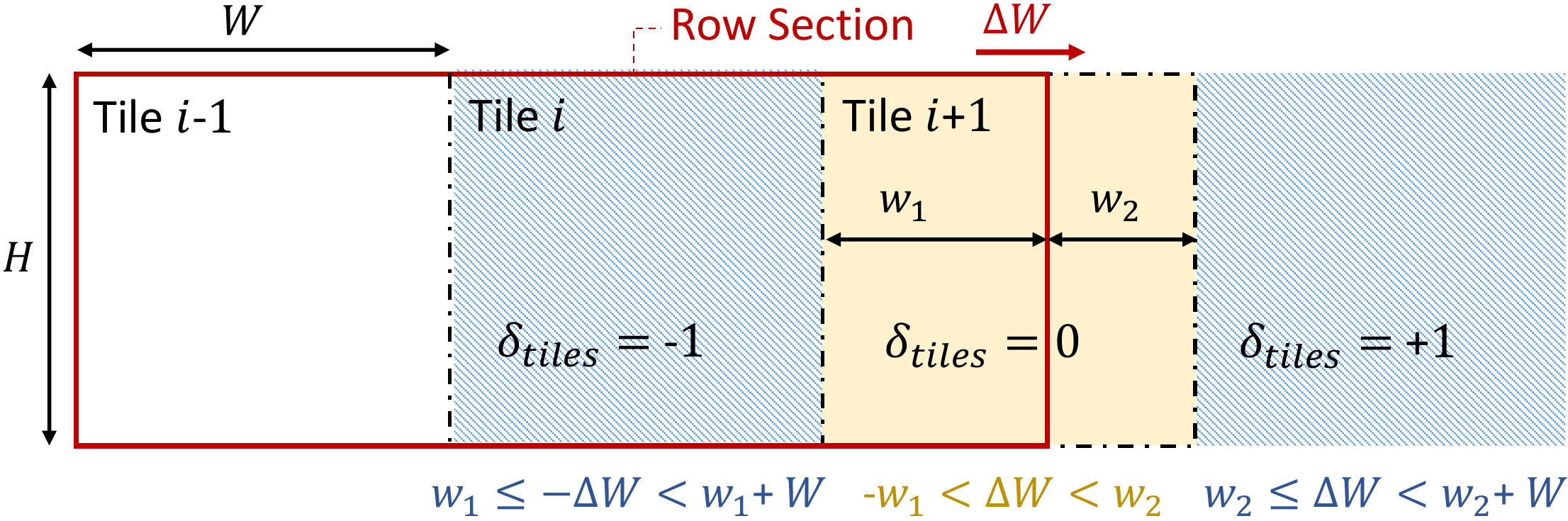} 	
	\caption{The Energy Difference ($E'-E$) for SA.}
	\label{fig: delta_energy}
\end{figure}

The pseudo code of the SA-based weight permutation algorithm is shown in Algorithm~\ref{alg_sa}. The optimization starts from a high temperature $T_{init}$ where any proposed solution is likely to be accepted. The subroutine $neighbor$-$state$ is invoked in Line 6 to propose a neighbor state ($state'$) by a one-step random permutation (either row or column). Since the optimal orders of rows and columns are correlated, it is important to mix the row swapping and column permutation during the optimization, instead of optimizing one first. Then, the $state'$ is packed and evaluated in Line 7 and 8. The energy difference in Line 8 is calculated by:
\begin{equation}\label{eq: energy_difference}
\begin{split}
 E'-E &= \delta_{size}+(\delta_{tiles}\times H\times W) \\
 \delta_{size} &= H\times\Delta W \\
\end{split}
\end{equation}
where the first term ($\delta_{size}$) is the difference in the matrix size after packing, and the second term is the difference in the number of weight tiles ($\delta_{tiles}$) multiplied by the size of the systolic array ($H\times W$).
An example is shown in Fig.~\ref{fig: delta_energy}. Since a slight change in the matrix size may not result in a difference in the number of weight tiles to map to the systolic array, $\delta_{tiles}$ will be zero in most cases. Once a relatively large difference is made, or the change is at the boundary of a tile (which reduces or increases the number of weight tiles), extra reward ($\delta_{tiles}<0$) or penalty ($\delta_{tiles}>0$) will be given. If $E'-E<0$, the new state is better and will always be accepted. Otherwise, it will have a probability of $e^{-\frac{E'-E}{temp}}$ to get accepted.
At high temperatures, the optimizer has a larger chance to accept worse solutions, whereas it will become more conservative as the temperature decreases.
The temperature is multiplied by $(1-f)$ ($f\in(0,1)$) after every $iter$ steps to simulate the cooling process. In this work, $f$ and $iter$ are set to 0.01 and 15, respectively. The empirical value of $T_{init}$ is from 1000 to 3000, depending on the size of the weight matrix. For the early layers with a small size (\textit{e.g.} $64\times64$), an initial temperature of 1000 is enough. For large layers, a higher temperature is needed. $T_{end}$ is set to $10^{-5}$ for all layers. 

\subsection{Hardware Design}
\label{subsec: hardware}

\begin{figure}[!t]
\setlength{\belowcaptionskip}{-14pt}
\centering
	\includegraphics[width=0.485\textwidth]{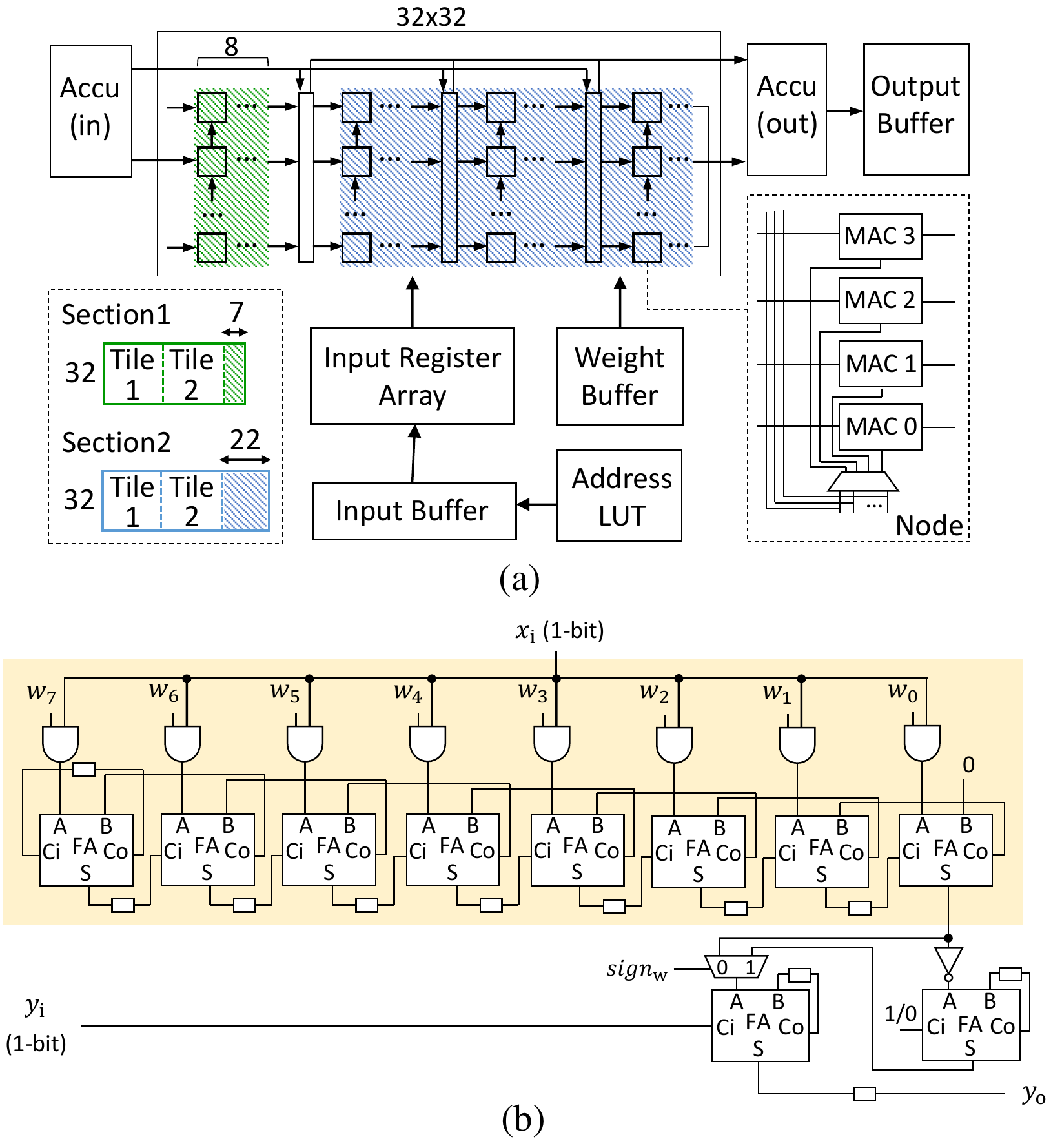} 	
	\caption{(a) The Systolic Array Architecture and (b) a MAC Unit.}
	\label{fig: archi}
\end{figure}

The hardware architecture to implement the compressed models is shown in Fig.~\ref{fig: archi}~(a). We adopt the weight-stationary flow described in Section \ref{subsec: systolic_array}. The overall systolic array architecture is similar to that of \cite{PackCNN} (in which bit-serial computation is used) with the following differences. Before computation, one or more row sections are loaded from the off-chip memory to the weight buffer.
Assume the size of the systolic array is $32\times32$. To perform matrix multiplication, a $32\times32$ weight tile in the order of the permutated and packed matrix is read from the buffer and stored in the nodes of the systolic array. Each column in the tile can pack up to 16 sparse columns of the original weight matrix. Therefore, the input activations also need to be grouped accordingly before sent to the systolic array. Since the order of the columns and the group composition are different for each row section, the corresponding input activations have to be read from the input buffer and aligned accordingly. The indices of the input channels corresponding to each packed column are known from the compression results and are stored in an address look up table (LUT). According to the index, the data of input channel is read from the input buffer and stored in the register array in a way that aligned with the column packing and order.
We use double buffering to prefetch the input data for the 32 systolic array columns to make the systolic array fully occupied during the computations. 

The matrix multiplication is performed in a bit-serial manner similar to \cite{PackCNN}. At each cycle, 1-bit of each input activation is shifted into the array. The 16 input bits (16 inputs that are packed to the same column) will propagate along the systolic array column, and the input data of adjacent columns have a one-cycle skew. At each node, 1-bit ($x_i$) is selected from the 16 input bits to multiply with the 8-bit weight. The partial sum will be accumulated along the row and sent to the output buffer. There is a one-cycle skew between the partial sums of adjacent rows. Similar to \cite{PackCNN}, we adopt an 8-bit representation for the weights and activations and a 32-bit representation for the partial sums. Since it takes 8 cycles to shift in the 8-bit activations and 32 cycles to accumulate the partial sums, each node has 4 MAC units to receive the input data in an interleaved manner to maximize the throughput. In each MAC unit (Fig.~\ref{fig: archi}~(b)), the main part for the bit-serial multiplication contains 8 full adders (FAs). The 4 MAC units share the same weight and index information.

The $32\times32$ systolic array is composed of four $32\times8$ subarrays that can be configured to work in two independent groups to maximize the hardware utilization. In the case that a row section has too few column groups to fill up the entire array, it can be folded with another row section and mapped together to the array to save the overall execution time and energy consumption. For example, after computing the first two tiles of row section 1 in Fig.~\ref{fig: archi}~(a), the last 7 column groups can be mapped to the first subarray, and the other subarrays can be mapped with the column groups of row section 2.
After the weight tile finishes computing with the data in the input buffer, the next tile will be loaded into the systolic array for computation. This process will continue until all the tiles in the weight buffer finish the computations with the data in the input buffer. Then, new data will be loaded to the input buffer to continue the computations. Once all the computations related to the current weight tiles are completed, the weight buffer can be updated to compute other output channels. This flow guarantees that each weight in the model will only be read once from the off-chip memory. Moreover, since the input buffer will finish computing with the entire row section before loading new data, all the input channels loaded on-chip can be fully utilized for computations.

\section{Subword-Level Tight Compression: Towards Higher Compression Rate}
\label{sec: sw_compress}
In Section~\ref{sec: tight_compression}, we have proposed a weight-level compression method. After compression, the model is uniformly quantized to 8-bit for the deployment in the systolic array system.
In this section, we will propose a subword-level compression method which leverages the mixed-precision representation to further enhance compression performance and reduce computational complexity. The compressed mixed-precision weight matrix can be easily supported by the systolic array with minor modification in the MAC units.

\begin{figure}[!t]
\centering
	\includegraphics[width=0.485\textwidth]{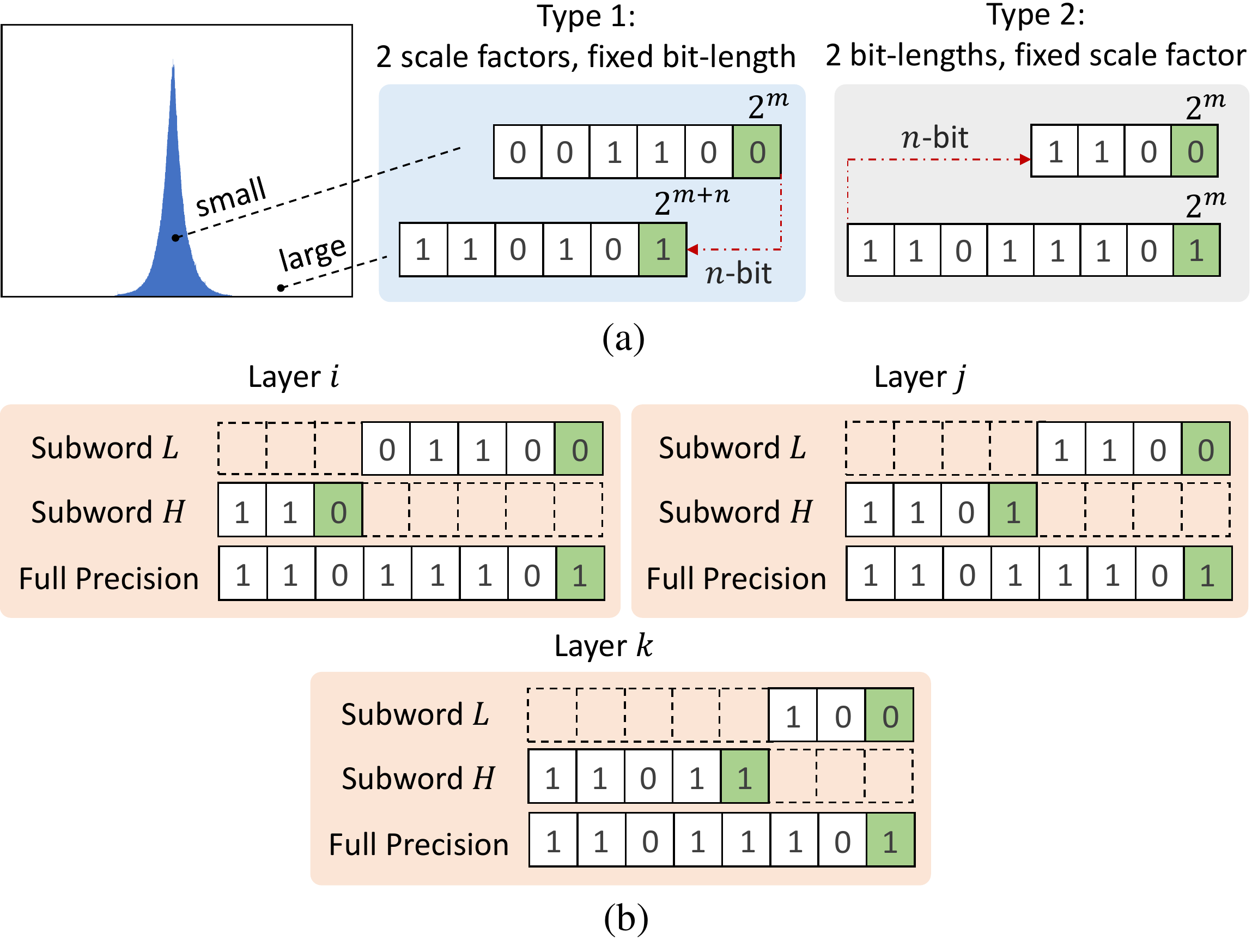} 	
	\caption{Mixed-Precision Weights (a) in Previous Works \cite{BiScale, OutlierAware} and (b) after the Proposed Subword Pruning.}
	\label{fig: sw_prune}
\end{figure}

\subsection{Subword Pruning}
For the subword-level tight compression, there is an extra step between the training stage (with unstructured weight pruning) and the compression stage (with weight permutation), where the sparse weight matrix of each layer is quantized to 8-bit and further pruned at the subword level.
The intuition of subword pruning is to represent the sparse weight matrix using mixed precisions. In Section \ref{subsec: related_works}, two types of mixed-precision representation are introduced, as shown in Fig.~\ref{fig: sw_prune}~(a). The first type is to use the same bit-length but two different scale factors to represent the weights. The second type is to use the same scale factor but two different bit-lengths.
Unlike the previous works \cite{BiScale, OutlierAware}, in subword pruning, both the bit-length and scale factor will be changed for the weights according to their magnitude. Fig.~\ref{fig: sw_prune}(b) illustrates this mixed-precision representation. The nonzero weights at each layer are in three formats after the subword pruning:  the \textbf{Subword \textit{L}}, the \textbf{Subword \textit{H}}, and the \textbf{8-bit full precision}.
For the weights with small magnitude, the most significant bits (\textit{i.e.} Subword $H$) in the 8-bit representation are zero and thus can be removed without incurring accuracy loss. On the contrary, for the weights with large magnitude, it is important to preserve a large range but may not be necessary to have such high resolution. In this case, the least significant bits (\textit{i.e.} Subword $L$) in the 8-bit representation can be removed with negligible accuracy loss. This operation can be interpreted as a fine-grained subword-level pruning, where either the Subword $L$ or Subword $H$ is pruned for most nonzero weights. A small fraction of weights is allowed to preserve the 8-bit full precision to limit the accuracy loss. Specifically, if the deviation after the subword pruning ($\Delta = |\frac{Subword_H - W_{8b}}{W_{8b}}|$) is larger than a threshold value (\textit{e.g.} 25\%), full precision will be used for the weight. A larger threshold value will result in less full-precision weights, which is beneficial for the subword-level compression. However, the accuracy loss induced by subword pruning will increase at the same time. The neural network needs to be retrained with subword pruning for several epochs to recover the accuracy.

\begin{figure}[!t]
\setlength{\belowcaptionskip}{-8pt}
\centering
	\includegraphics[width=0.485\textwidth]{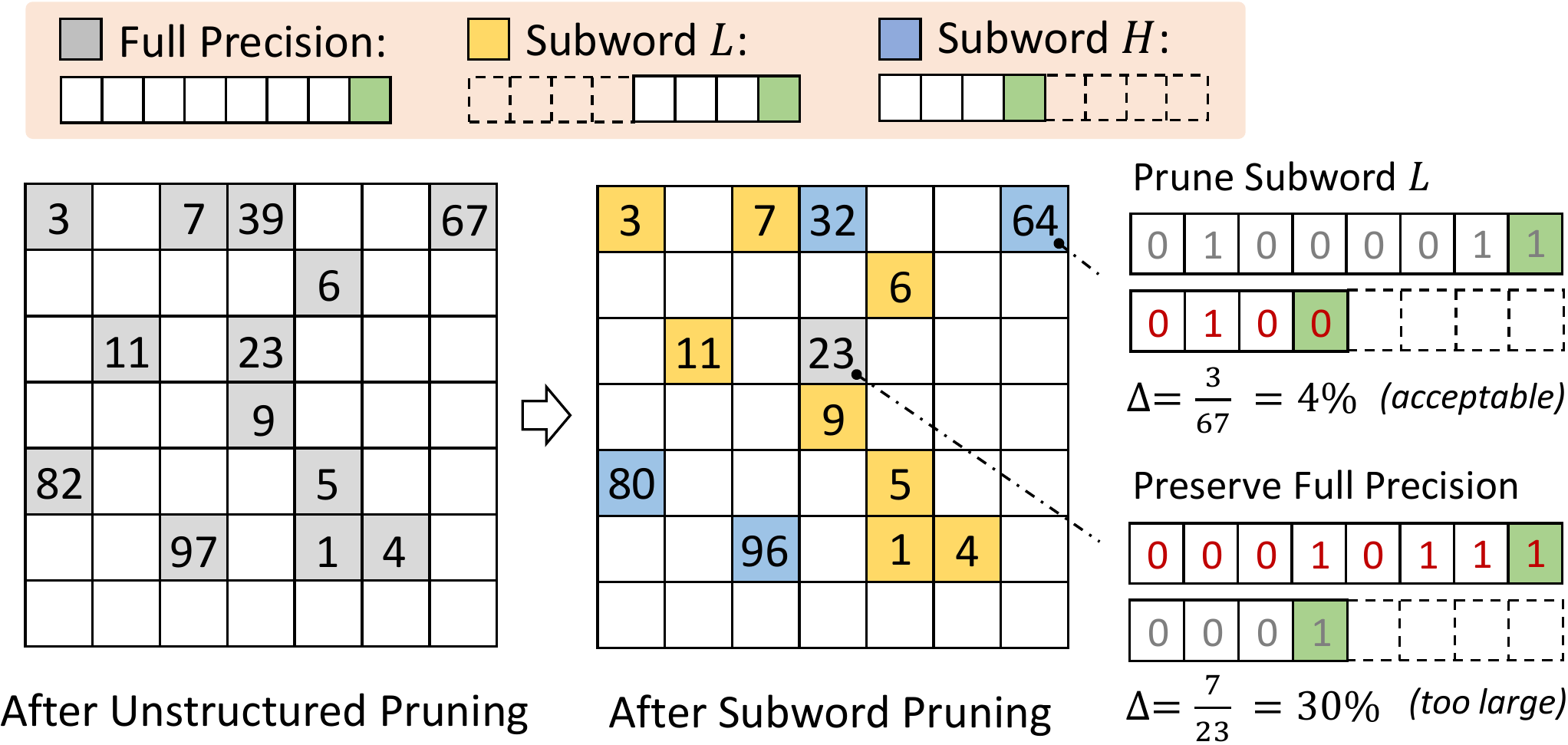} 	
	\caption{Weight Matrix after the Subword Pruning.}
	\label{fig: matrix_after_sw_prune}
\end{figure}

An illustrating example is shown in Fig.~\ref{fig: matrix_after_sw_prune}. In this example, the bit-length is 4-bit for the subwords, and the allowable deviation for subword pruning is set to 25\%. Most nonzero weights only have one nonzero subword after the pruning. A small fraction of weights (\textit{e.g.} the one with a value of 23) preserves 8-bit precision. Later, this fine-grained subword-level sparsity will be exploited to improve the compression performance. It is worth mentioning that the bit-lengths of the Subword $H$ and Subword $L$ may vary over different layers but are consistent within a layer for efficient model compression. Details for determining the bit-lengths will be discussed shortly.

\subsection{Exploiting the Subword Sparsity for Compression}
\label{subsec: exploit_sw_sparsity}

\begin{figure}[!t]
\setlength{\belowcaptionskip}{-8pt}
\centering
	\includegraphics[width=0.485\textwidth]{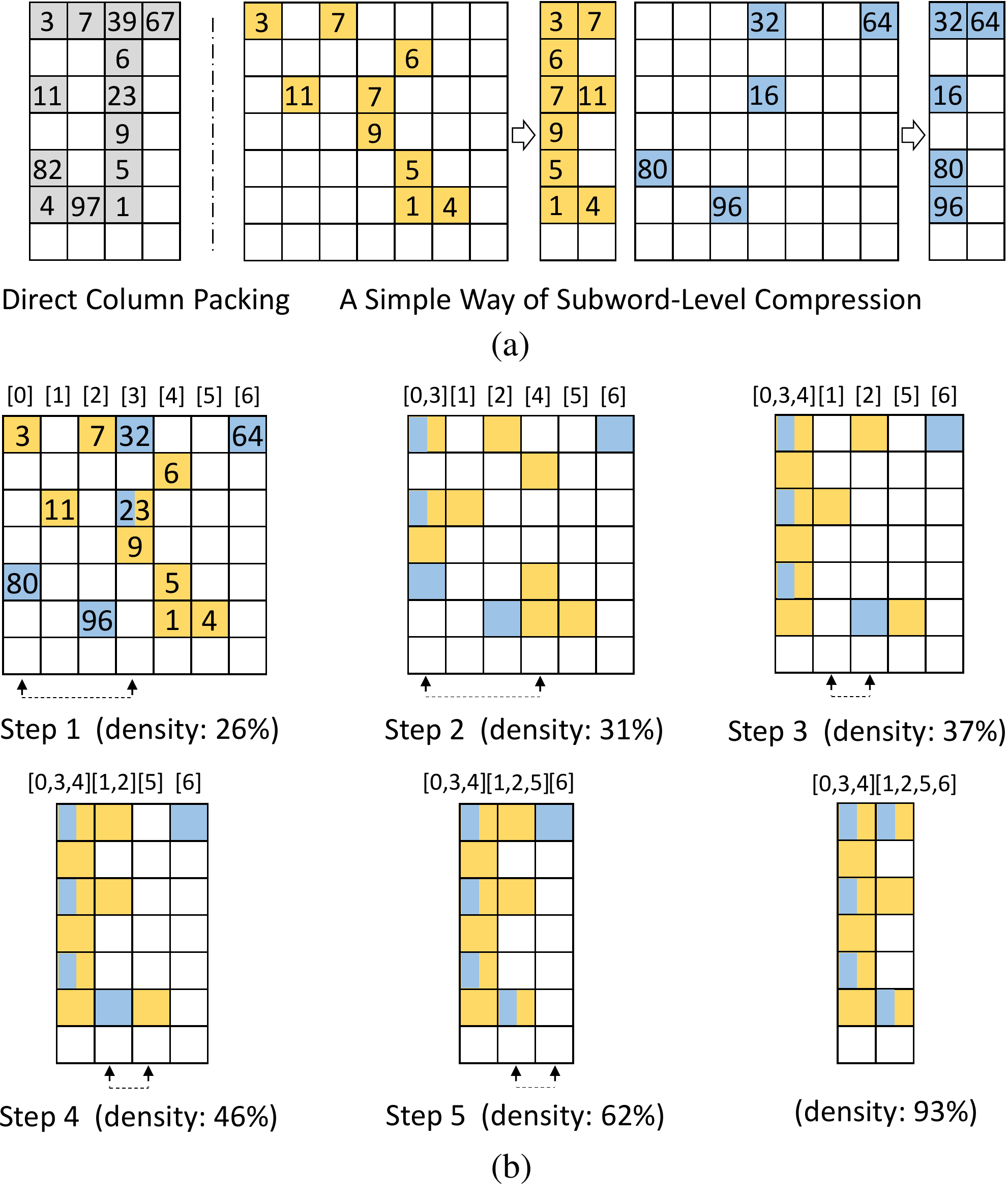} 	
	\caption{(a) A Simple Way of Subword-Level Compression and (b) the Proposed Subword-Level Compression.}
	\label{fig: sw_compression}
\end{figure}

To illustrate the subword-level compression, the small weight matrix shown in Fig.~\ref{fig: matrix_after_sw_prune} is firstly compressed by column packing. (Weight permutation is not applied here for simplicity.) The compressed weight matrix is shown in Fig.~\ref{fig: sw_compression}~(a). In a column group, only one nonzero weight is allowed at each row position. Any conflicts among the columns will make the packing invalid.
A simple way to exploit the subword sparsity for compression is to divide the original weight matrix into two matrices, one for Subword $H$ and the other for Subword $L$. Then, each subword matrix can be compressed individually, as shown in Fig.~\ref{fig: sw_compression}~(a). Although the overall matrix size after compression ($7\times2\times2$) is the same as the original column packing, each node of the systolic array can be less complex due to the smaller bit-length of the subwords.
However, this naive compression method will bring some problems. Since a column in the original weight matrix is now split into two columns, each input activation needs to be grouped and input twice into the systolic array to finish the computations. It will increase the energy consumption for input preparation and hence induce non-trivial overhead.
Besides, the address LUT (Fig.~\ref{fig: archi}) that stores the indices for aligning the input activations will be doubled since the column group compositions for the two subword matrices are different. 

A more efficient compression method is shown in Fig.~\ref{fig: sw_compression}~(b), where the entire weight matrix is compressed as a whole. Two nonzero weights at the same row are allowed to be merged during the compression, as long as their nonzero subwords are not in the same position. Later, the merged weights will be mapped to the same node of the systolic array and computed as a whole.
The density of the weight matrix in Fig.~\ref{fig: sw_compression}~(b) is calculated by the number of nonzero weights divided by the size of the compressed weight matrix. Originally, the weight matrix has a small density of 26\%.
In the first step, column~3 is selected and grouped with column~0. The two weights at the first row are merged and become 8-bit \{Subword $H$, Subword $L$\}. After grouping the two columns, the density increases to 31\%. This process will continue until the first column group can no longer combine with any columns due to conflicts or reaching the upper bound of columns per group (parameter $G$ in Algorithm~\ref{alg_col_packing}). Then the next column group will be processed. After compression, the size of the weight matrix is reduced to $7\times2$, and the density increases from 26\% to 93\%.
Compared to the previous solution in Fig.~\ref{fig: sw_compression}~(a), no 
extra overhead will be caused by input preparation or address LUT.
This example shows how subword sparsity can help to improve the compression performance. It can be combined with the simulated annealing-based weight permutation proposed in Section~\ref{sec: tight_compression} for a complete subword-level compression.

To obtain the optimal compression performance, it is preferred to have half of the nonzero weights pruned to Subword $L$ and the other half pruned to Subword $H$.
In this case, the nonzeros at the same row will have a larger chance to be merged to avoid conflicts. The ratio of Subword $H$ to Subword $L$ depends on how we partition the 8-bit weight. For example, Table~\ref{tab: sw_bit-length} shows the bit-lengths of the subwords and the proportion of each subword type at a layer of the benchmark on CIFAR-10 \cite{Cifar10}. The optimal bit-length is 4-bit for both subword types in this example.
In the experiment, three optimal combinations are observed for the layers of all the benchmarks, including \{3-bit $H$, 5-bit $L$\}, \{4-bit $H$, 4-bit $L$\}, and \{5-bit $H$, 3-bit $L$\} (Fig.~\ref{fig: sw_prune}~(b)). The hardware can be configured to support different combinations.

\begin{table}[tb]
\setlength{\abovecaptionskip}{2pt}
\centering
\renewcommand{\arraystretch}{1.25}
\caption{Bit-Lengths of the Subwords and the Proportion of Each Subword Type at a Layer (CIFAR-10)}
\begin{center}
\begin{tabular}{|c|c|c|c|c|c|}
\hline
\multirow{2}{*}{Nonzero Weights}
& \multicolumn{5}{c|}{Bit-Lengths \{$H$,$L$\}} \\
\cline{2-6}
& \{2,6\}
& \{3,5\}
& \{4,4\}
& \{5,3\}
& \{6,2\} \\
\hline
Subword $L$ (\%)
& 99.9 
& 97.6 
& 57.7 
& 18.4 
& 6.0 
\\
\hline
Subword $H$ (\%)
& $<$0.1 
& 2.3 
& 37.6 
& 76.4 
& 90.7 
\\
\hline
8-Bit Full Precision (\%)
& 0
& 0.1 
& 4.7 
& 5.2 
& 3.3 
\\
\hline
\end{tabular}
\label{tab: sw_bit-length}
\end{center}
\vspace{-10pt}
\end{table}

\subsection{Hardware Modification for Subword-Level Compression}
\label{subsec: sw_hardware}
\begin{figure}[!t]
\setlength{\belowcaptionskip}{-6pt}
\centering
	\includegraphics[width=0.485\textwidth]{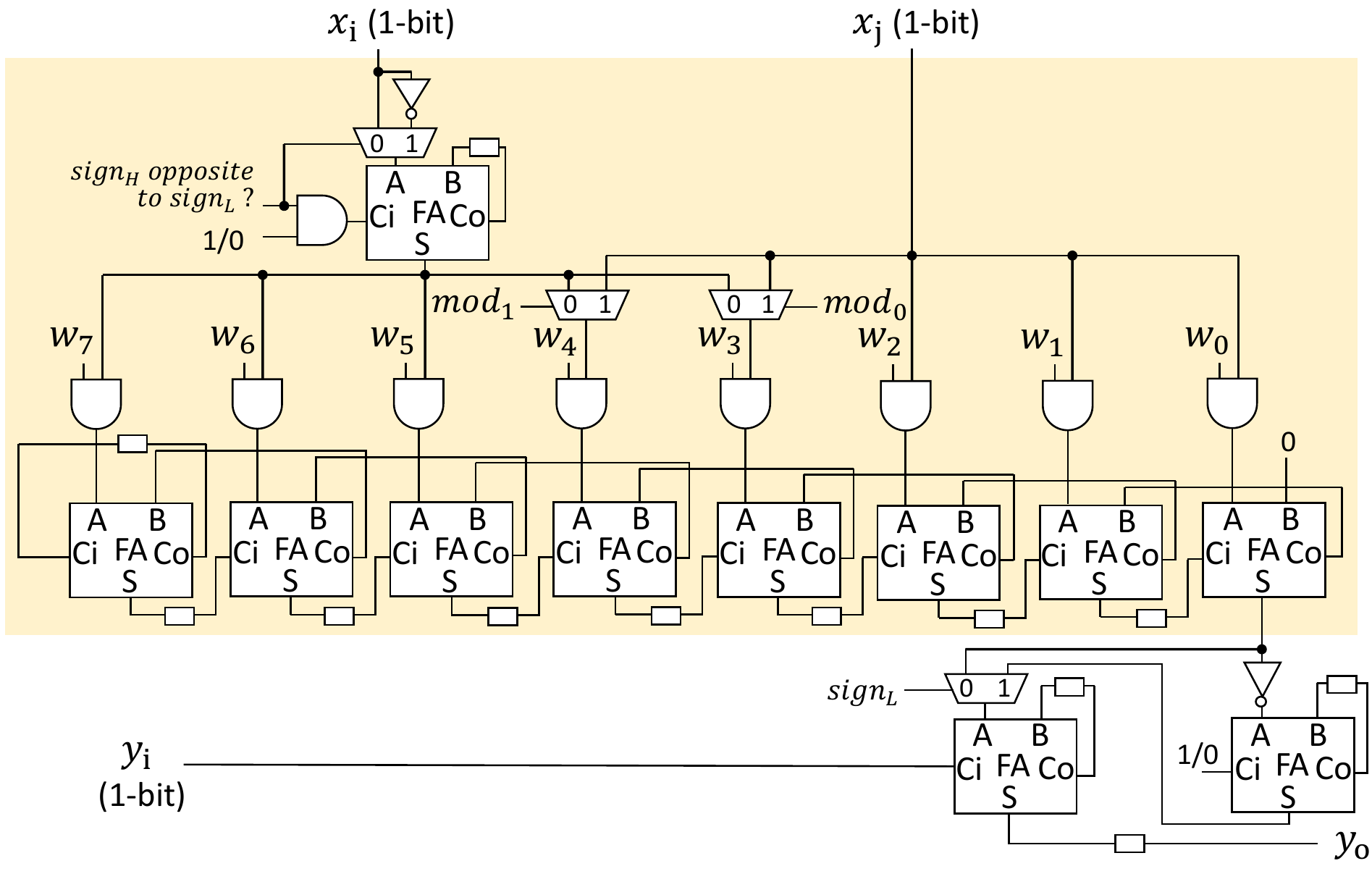} 	
	\caption{The MAC Unit for Subword-Level Compression.}
	\label{fig: sw_mac}
\end{figure}

The top-level hardware architecture for implementing the subword-level compressed model is similar to that in Fig.~\ref{fig: archi}~(a). Each column group in the weight tile can pack up to 16 sparse columns of the original weight matrix, and the input preparation part is the same as before.
One major difference from the previous design is the MAC unit at each node of the systolic array. Fig.~\ref{fig: sw_mac} shows the MAC unit modified to support the subword-level compression. Each nonzero entry of the weight tile can be a Subword $H$, a Subword $L$, or an 8-bit \{Subword $H$, Subword $L$\} in the case that two weights are merged or the full precision is used for one weight. Therefore, up to 2 bits ($x_i$ and $x_j$) are selected to multiply with the subwords stored at each node. An extra full adder is used to invert the input activation for Subword $H$ if the two subwords have opposite signs. As mentioned in Section~\ref{subsec: exploit_sw_sparsity}, there are three bit-length combinations for the subwords. It is supported by the MAC unit and controlled by the 2-bit signal $mod$.
Besides, extra index information needs to be stored for the compressed weight matrix. Originally, a 4-bit index is needed at each node to select 1-bit input from the 16 input bits. After subword-level compression, two sets of the 4-bit index are needed, one for Subword $H$ and the other for Subword $L$. In the case that an 8-bit full-precision weight is stored at the node, the two indices will have the same value to select the same input bit ($x_i$=$x_j$).
Although energy and area overhead is induced by the modification, the overall energy efficiency and area efficiency are improved due to better compression performance. A detailed analysis will be given in the experimental results.

\section{Experimental Results}
In this section, we will evaluate the performance of the proposed compression method and compare it with the previous compression techniques. The weight-level compression results are presented first in Section~\ref{subsection: compression_performance_1}. Then, the models are further pruned and compressed at the subword level to improve the compression results in Section~\ref{subsection: compression_performance_2}.
After that, we implement and synthesize the hardware architectures presented in 
the previous sections,
and the throughput and energy efficiency of the implementations are reported in Section~\ref{subsection: implement_performance}.
The buffers are implemented by SRAM and modeled in CACTI 7.0 \cite{cacti70} using 45nm process node to estimate the energy and area. Other components, including the systolic array, the accumulation units, and the input register array, are synthesized using the Synopsys Design Compiler with Nangate 45nm Open Cell Library \cite{nangate}.
To make a comparison with \cite{PackCNN}, we also implement the models compressed by conflict pruning in a baseline systolic array system similar to \cite{PackCNN}.

\subsection{Benchmarks}
We evaluate the compression method using three popular datasets, including the CIFAR-10 dataset \cite{Cifar10} with $32\times32$ color images of 10 classes, the CIFAR-100 dataset \cite{Cifar10} with $32\times32$ images of 100 classes, and the ImageNet dataset \cite{ImageNet} with $224\times224$ images of 1000 classes. We adopt the same network architecture as \cite{PackCNN} to make a fair comparison with the conflict pruning method. The neural networks are composed of 19 convolutional layers and 1 fully connected layer \cite{ShiftNet}. Stochastic gradient descent is used for training. Similar to \cite{PackCNN}, the Nesterov momentum is set to 0.9. The learning rate is initialized as 0.1 and gradually decays to 0. L1 regularization is used with the strength equal to $10^{-7}$.
The neural networks on the three datasets are trained for 300, 300, and 120 epochs, respectively, with a batch size of 128. After training, the models with unstructured weight sparsity are obtained.
For subword pruning, the models are further retrained for 20, 20, and 8 epochs, respectively, to recover the accuracy.

\subsection{Performance of Weight-Level Tight Compression}
\label{subsection: compression_performance_1}
\begin{figure}[!t]
\setlength{\belowcaptionskip}{-6pt}
\centering
	\includegraphics[width=0.49\textwidth]{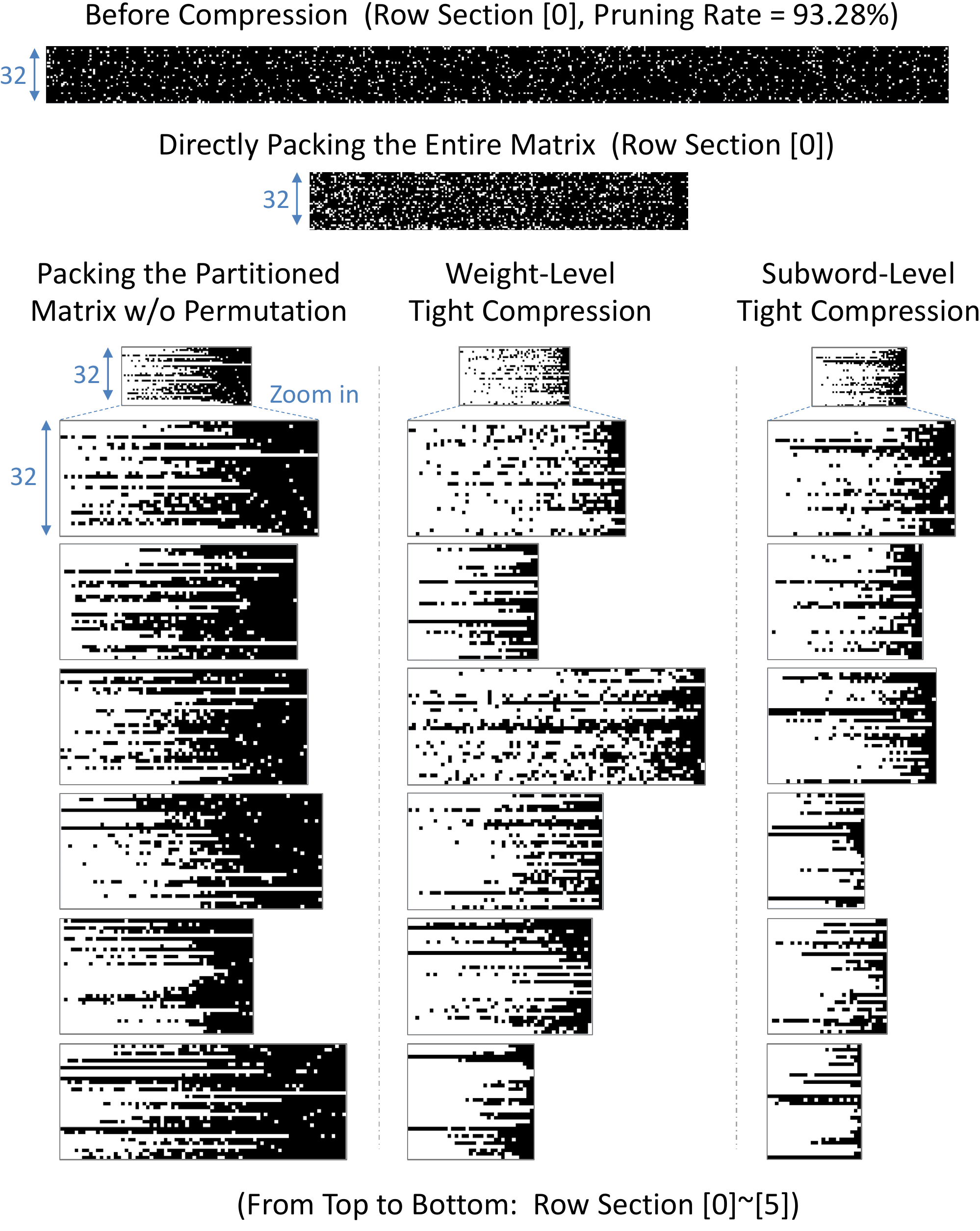} 	
	\caption{Weight Matrix Compressed Using Different Methods (Black: Zero Weights, White: Nonzero Weights).}
	\label{fig: compress_performance}
\end{figure}

The initial number of weights without pruning is 1.91M in the benchmark of CIFAR-10. During training, 93.3\% of the weights can be pruned with a negligible accuracy loss. The top-1 accuracy after pruning is 93.10\%.
Similarly, the benchmarks of CIFAR-100 and ImageNet initially have 3.33M and 4.49M weights, respectively. The pruning rates in the two benchmarks are 90.4\% and 88.6\%, respectively. After pruning, 75.09\% top-1 accuracy is obtained on CIFAR-100, and 56.616\% top-1 accuracy is obtained on ImageNet.
Fig.~\ref{fig: compress_performance} shows the row sections of a typical weight matrix ($512\times512$, CONV15) in the benchmark of CIFAR-10 before and after compression. Each row section has a size of $32\times512$. 93.3\% of weights in the matrix are pruned and represented by the black dots. Other weights are represented by the white dots.  

Directly packing the columns of the entire weight matrix without conflict pruning can only achieve a limited compression rate of 2.4$\times$. The first row section after column packing is shown in Fig.~\ref{fig: compress_performance}. The compressed weight matrix has a size of $512\times212$, and only 17\% of weights are nonzero. In this case, the high sparsity cannot be fully utilized to improve energy efficiency and throughput.
After partitioning the weight matrix and packing each row section independently (without permutation), a better compression result with 53\% of nonzero weight density is obtained. Fig.~\ref{fig: compress_performance} shows the row section [0]$-$[5] in the compressed weight matrix.
The compression density can be further improved by weight permutation. Through the weight-level tight compression, the 512 columns in the first row section can be packed in 61 groups, thus effectively compressing the row section by 8.4 times. 78\% of weights are nonzero in the compressed row section, which is much denser compared to the original sparse format. Similar improvement has been observed for other row sections, as shown in Fig.~\ref{fig: compress_performance}.
The compressed weight matrix tends to have denser column groups on the left. It is because the packing starts from the left and searches the remaining columns to form the densest group at each step. No accuracy loss is caused by the weight permutation. 

\begin{table*}[tb]
\setlength{\abovecaptionskip}{4pt}
\centering
\renewcommand{\arraystretch}{1.2}
\caption{Performance of Tight Compression and the Comparison with Conflict Pruning \cite{PackCNN}}
\begin{center}
\begin{tabular}{|c|c|c|c|c|}
\hline
\textbf{Benchmark}
& \textbf{Performance}
& \textbf{Conflict Pruning \cite{PackCNN}}
& \textbf{Weight-Level Tight Compression}
& \textbf{Subword-Level Tight Compression}\\
\hline
\multirow{4}{*}{CIFAR-10}  & Pruning Rate (\%)   & 84.7  & 93.3	  & 93.3 \\
\cline{2-5} 
	& Number of Nonzero Weights  & 0.3M  & 0.13M  & ~~~0.13M ~~(4.64\% in Full-Precision)~ \\
\cline{2-5} 
  	& Matrix Compression Rate  & 5.88  & 10.28  & 14.13\\
\cline{2-5} 
	& Top-1 Accuracy (\%)  & 92.9  & 93.1	  & 92.16 \\
\hline
\hline
\multirow{4}{*}{CIFAR-100}	& Pruning Rate (\%)  & 80.5  & 90.4  & 90.4 \\
\cline{2-5} 
	& Number of Nonzero Weights  & 0.65M  & 0.32M  & ~~~0.32M ~~(10.33\% in Full-Precision)~ \\
\cline{2-5} 
	& Matrix Compression Rate  & 4.60  & 8.30  & 11.10 \\
\cline{2-5} 
	& Top-1 Accuracy (\%)  & 75.08  & 75.09  & 74.32 \\
\hline
\hline
\multirow{4}{*}{ImageNet}& Pruning Rate (\%)  & 66.6  & 88.6  & 88.6 \\
\cline{2-5} 
	& Number of Nonzero Weights  & 1.5M  & 0.51M  & ~~~0.51M ~~(17.90\% in Full-Precision)~ \\
\cline{2-5} 
	& Matrix Compression Rate  & 2.69  & 6.52  & 8.66 \\
\cline{2-5} 
	& Top-1 Accuracy (\%)  & 55.99  & 56.61  & 55.70 \\
\hline
\end{tabular}
\label{tab: compress_performance}
\end{center}
\vspace{-6pt}
\end{table*}

The pruning rate, matrix compression rate, and accuracy after compression are summarized in Table~\ref{tab: compress_performance}. Compared to the conflict pruning method \cite{PackCNN}, the weight-level tight compression can achieve a larger pruning rate. It is because accuracy loss will be introduced in conflict pruning and hence to achieve the same level of accuracy, more weights have to be kept.
The number of nonzero weights after tight compression is 2.3 times smaller than conflict pruning in the benchmark of CIFAR-10. The matrix compression rate is effectively improved from 5.88$\times$ to 10.28$\times$, and thus significant improvements can be achieved in throughput and energy efficiency (which will be shown in Section~\ref{subsection: implement_performance}).
Similarly, on the CIFAR-100 dataset, the pruning rate in tight compression can be 10\% larger than the conflict pruning method when the two models are of the same accuracy. As a result, the number of nonzero weights after tight compression is 2.0 times smaller than the conflict pruning method. The matrix compression rate can be improved from 4.60$\times$ to 8.30$\times$ by the weight-level tight compression.
On the ImageNet dataset, the number of nonzero weights after tight compression is 2.9 times smaller than the conflict pruning method, and the matrix compression rate can be improved from 2.69$\times$ to 6.52$\times$.

We also compare tight compression with some structured pruning methods in terms of the model size and accuracy. The results are shown in Table~\ref{tab: structured_prune_comparison}. Compared to structured pruning, tight compression can achieve a smaller matrix size with the same level of accuracy.

\begin{table}[tb]
\setlength{\abovecaptionskip}{2pt}
\centering
\renewcommand{\arraystretch}{1.2}
\caption{Comparison between Tight Compression and Structured Pruning on CIFAR-10}
\begin{center}
\begin{tabular}{|>{\centering\arraybackslash} m{2.05cm}|>{\centering\arraybackslash} m{1.1cm}|>{\centering\arraybackslash} m{0.89cm}|>{\centering\arraybackslash} m{0.89cm}|>{\centering\arraybackslash} m{1.6cm}|}
\hline
\multirow{3}{*}{\textbf{Performance}} 
	& \textbf{Network} 
	& \textbf{Filter} 
	& \textbf{Filter} 
	& \textbf{Tight} \\
	& \textbf{Slimming} 
	& \textbf{Pruning}
	& \textbf{Pruning}
	& \textbf{Compression} \\
	& \cite{NetworkSlimming}  & \cite{FilterPruning1}  & \cite{FilterPruning2}  &  \textbf{\scriptsize{(Weight-Level)}} \\
\hline
Pruning Rate (\%)  & 88.5  & 64.0  & 86.5  & 93.3  \\
\hline
Weight Matrix Size after Compression  & 2.30M  & 5.4M  & 2.02M  & 0.18M \\
\hline
Top-1 Accuracy (\%)  & 93.8  & 93.4  & 90.9	 & 93.1 \\
\hline
\end{tabular}
\label{tab: structured_prune_comparison}
\end{center}
\vspace{-9pt}
\end{table}

\subsection{Performance of Subword-Level Tight Compression}
\label{subsection: compression_performance_2}
For better compression performance, the sparse model is further pruned and compressed at the subword level using the method presented in Section~\ref{sec: sw_compress}. The maximum allowable deviation ($\Delta_{max}$) for the subword pruning is set to 0.3 for the benchmark of CIFAR-10.
The weight pruning rate is the same as before (\textit{i.e.} 93.3\%). However, 95.36\% of the nonzero weights only have one subword (either Subword $L$ or Subword $H$). A small fraction of the nonzero weights (4.64\%) preserves 8-bit precision to limit the accuracy loss. The relative accuracy loss induced by subword pruning is 0.94\%.
After the subword-level compression, the total size of the weight matrices is reduced from 1.91M to 0.135M, and a large matrix compression rate of 14.13$\times$ is achieved for the model.
The results are summarized in Table \ref{tab: compress_performance}. Compared to conflict pruning and the weight-level tight compression, the matrix compression rate can be significantly improved by the fine-grained subword-level compression.
Fig.~\ref{fig: compress_performance} shows the row section [0]$-$[5] in the compressed weight matrix (CONV15 in the benchmark of CIFAR-10). A large compression rate of $15.21\times$ is achieved at the layer, which is close to the upper bound $16\times$ (parameter $G$ in Algorithm~\ref{alg_col_packing}). The density of nonzeros in the weight matrix is improved from 78\% to 101\% ($>$100\% due to the merging of subwords), compared to the weight-level tight compression.

Similar improvement has been achieved in the benchmark of CIFAR-100. The maximum allowable deviation ($\Delta_{max}$) for the subword pruning is set to 0.27, and only 10.33\% of nonzero weights preserve an 8-bit full precision. Others only have one nonzero subword. The relative accuracy loss induced by subword pruning is 0.77\%. Compared to the weight-level tight compression, the total size of the weight matrices is reduced from 0.40M to 0.30M. Therefore, the compression rate of the weight matrix is improved from 8.30$\times$ to 11.10$\times$.
In the benchmark of ImageNet, $\Delta_{max}$ is set to 0.23, and 17.90\% of nonzero weights preserve full precision after subword pruning. Compared to the weight-level tight compression, the compression rate of the weight matrix is further improved from 6.52$\times$ to 8.66$\times$ at the cost of 0.91\% of relative accuracy loss.

\subsection{Analysis of Throughput and Energy Efficiency}
\label{subsection: implement_performance}
\begin{figure}[!t]
\setlength{\belowcaptionskip}{-6pt}
\centering
	\includegraphics[width=0.465\textwidth]{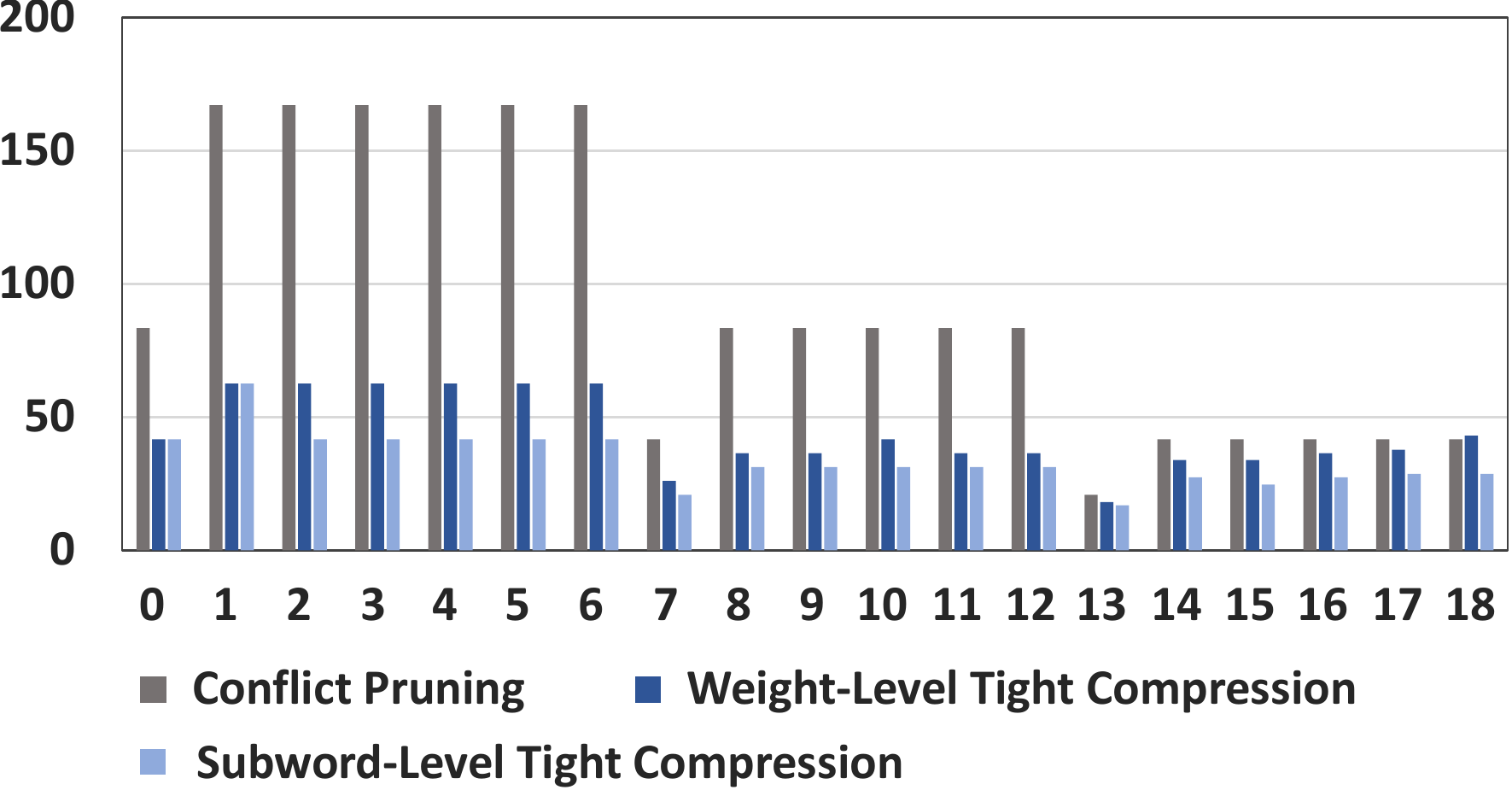} 	
	\caption{Execution Time of Each Layer in the Benchmark of CIFAR-10 (X Axis: Layer Index, Y Axis: Time (us)).}
	\label{fig: time}
\end{figure}

Since accuracy loss will be induced by conflict pruning, uneven pruning is performed in \cite{PackCNN} to maintain high accuracy while reducing the model size as much as possible.
Specifically, less aggressive pruning and column packing (1-4 columns per group) are performed in the early layers that are relatively small and have less capacity for compression, and more aggressive pruning is performed in the subsequent large layers.
Although the size of the large layers can be reduced, the improvement in throughput is limited since the first few layers have higher computational complexity (due to the large input volume) and take more time to process.
In the proposed tight compression method, uniform pruning is performed and thus all the layers can be compressed efficiently to improve the throughput and energy efficiency.
Fig.~\ref{fig: time} shows the execution time of each layer in the benchmark of CIFAR-10. Compared to conflict pruning, the overall execution time can be efficiently reduced through tight compression. Besides, Fig.~\ref{fig: time} also shows that the fine-grained subword-level tight compression can achieve a higher throughput than the weight-level tight compression.

\begin{table*}[tb]
\setlength{\abovecaptionskip}{4pt}
\centering
\renewcommand{\arraystretch}{1.2}
\caption{Throughput and Energy Efficiency of the Implementations for Tight Compression and Conflict Pruning}
\begin{center}
\begin{tabular}{|c|c|c|c|c|c|c|}
\hline
\multirow{2}{*}{\textbf{Benchmark}}
& \multirow{2}{*}{\textbf{Performance}}
& \textbf{Conflict Pruning}{\scriptsize \cite{PackCNN}}
& \multicolumn{2}{c|}{\textbf{Weight-Level Tight Compression}}  
& \multicolumn{2}{c|}{\textbf{Subword-Level Tight Compression}}
\\
\cline{4-7}
&
& \circled{1}
& ~~Value~~
& Improvement Over \circled{1}
& ~~Value~~
& Improvement Over \circled{1}\\
\hline
\multirow{4}{*}{CIFAR-10}
& Throughput (frames/s) 
& 563.2
& 1196.8 
& 2.12$\times$
& 1550.5 
& 2.75$\times$\\
\cline{2-7} 
& Energy Efficiency (frames/J)
& 2440.4
& 3827.4 
& 1.57$\times$
& 4537.4
& 1.86$\times$ \\
\cline{2-7} 
& Area Efficiency (frames/s/mm$^2$)	
& 332.3
& 602.9 
& 1.81$\times$
& 695.6 
& 2.09$\times$ \\
\cline{2-7} 
& Top-1 Accuracy (\%)
& 93.0
& 93.0
& 0
& 92.14
& -0.86\% \\
\hline\hline
\multirow{4}{*}{CIFAR-100}
& Throughput (frames/s) 
& 257.4
& 460.0
& 1.79$\times$
& 581.1
& 2.26$\times$ \\
\cline{2-7} 
& Energy Efficiency (frames/J)
& 1066.6
& 1441.7
& 1.35$\times$
& 1687.2
& 1.58$\times$ \\
\cline{2-7} 
& Area Efficiency (frames/s/mm$^2$)	
& 151.8
& 231.7
& 1.53$\times$
& 260.7
& 1.72$\times$ \\
\cline{2-7} 
& Top-1 Accuracy (\%)
& 74.80
& 75.06
& 0.26\%
& 74.17
& -0.63\% \\
\hline\hline
\multirow{4}{*}{ImageNet}
& Throughput (frames/s)
& 53.8
& 101.9
& 1.89$\times$	
& 133.4
& 2.48$\times$ \\
\cline{2-7} 
& Energy Efficiency (frames/J)
& 238.6
& 336.5 
& 1.41$\times$
& 405.0
& 1.70$\times$ \\
\cline{2-7} 
& Area Efficiency (frames/s/mm$^2$)	
& 31.8
& 51.3 
& 1.61$\times$
& 59.8
& 1.88$\times$ \\
\cline{2-7} 
& Top-1 Accuracy (\%)
& 51.28
& 55.86 
& 4.58\%
& 55.49
& 4.21\% \\
\hline
\end{tabular}
\label{tab: throughput_and_energy}
\end{center}
\vspace{-8pt}
\end{table*}

\begin{figure}[!t]
\setlength{\belowcaptionskip}{-6pt}
\centering
	\includegraphics[width=0.4\textwidth]{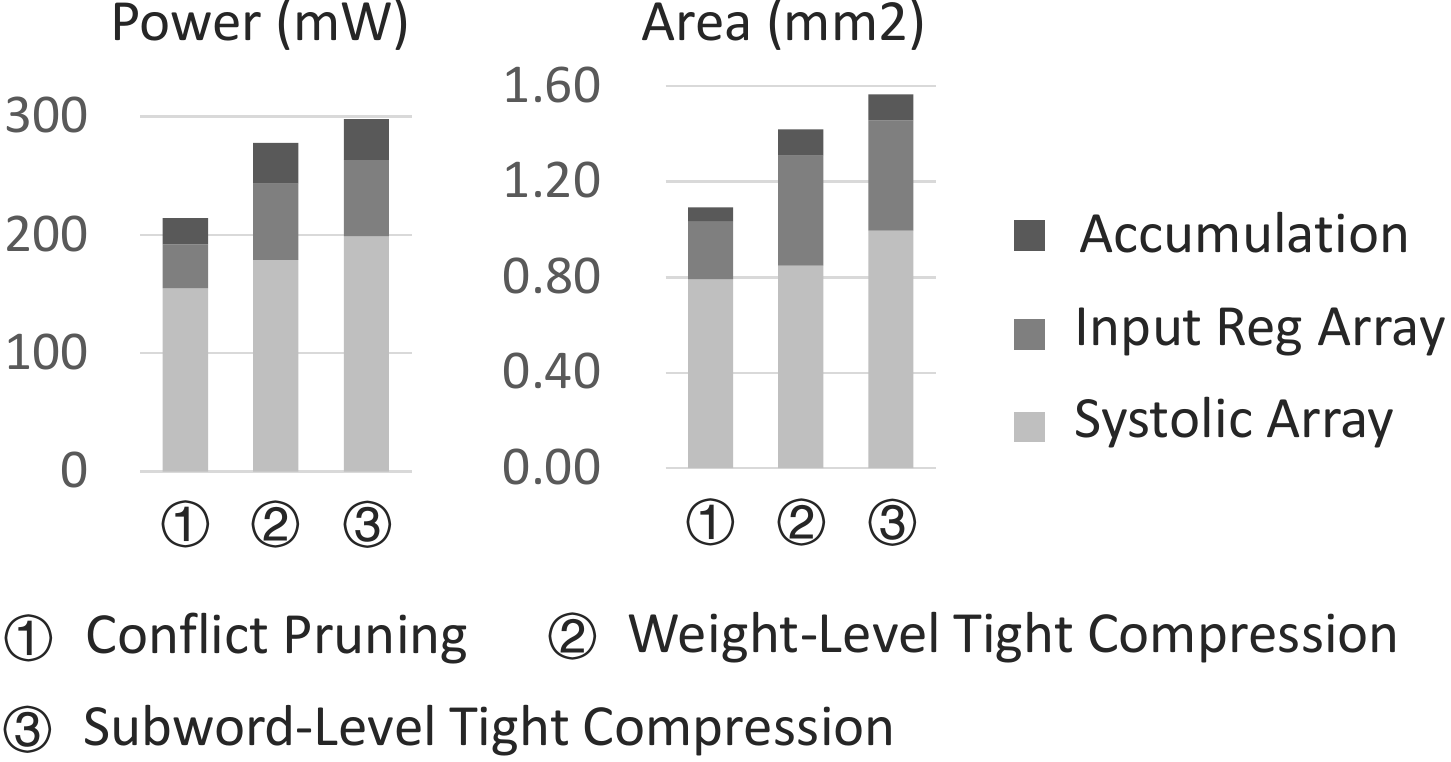} 	
	\caption{Synthesis Results for Different Implementations.}
	\label{fig: syn_results}
\end{figure}

The comparison of throughput, energy efficiency, and area efficiency among different compression methods are summarized in Table~\ref{tab: throughput_and_energy}. Compared to conflict pruning, the weight-level tight compression improves the throughput and energy efficiency by 2.12 and 1.57 times, respectively, on the CIFAR-10 dataset. Fig.~\ref{fig: syn_results} shows the synthesis results for different implementations. The power consumption and area of the systolic array are slightly larger than the baseline case. Since more columns can be packed in one group through tight compression, the input register array is larger than the baseline implementation for conflict pruning which only supports eight columns per group at maximum. Besides, the address LUT also has some area overhead. Therefore, the area of the implementation for the weight-level tight compression is 20\% larger than the baseline implementation for conflict pruning.
However, as the throughput is increased, the overall area efficiency is improved by 1.81 times.
In the benchmark of CIFAR-100, the throughput, energy efficiency, and area efficiency are improved by 1.79, 1.35, and 1.53 times, respectively.
Similarly, in the benchmark of ImageNet, the throughput, energy efficiency, and area efficiency are improved by 1.89, 1.41, and 1.61 times, respectively. The top-1 accuracies in the implementations for weight-level tight compression and conflict pruning are 55.86\% and 51.28\%, respectively, on the ImageNet dataset. The accuracies are different from those in Table~\ref{tab: compress_performance} since the activations are also quantized.

As discussed in Section~\ref{subsec: sw_hardware}, extra energy and area overhead is induced to support the subword-level compression. The overhead contains two parts. The first part is related to the modification in the MAC units at each node of the systolic array. The power and area of the systolic array are increased by 11\% and 17\%, respectively, compared to the implementation for the weight-level tight compression.
The second part of the overhead is related to the extra index information for the weight matrix. Since the energy consumption and area of the weight buffer are much smaller ($<$10\%) than the systolic array
, the overall overhead is mainly determined by the first part.
Since better compression performance is obtained through the subword-level compression, the overall throughput, energy efficiency, and area efficiency are significantly improved, as shown in Table~\ref{tab: throughput_and_energy}.

\section{Conclusions}
In this paper, we have proposed a compression method to fully utilize the unstructured sparsity for implementing neural networks efficiently on the regular systolic array architecture.
Two pruning granularities are explored for weight matrix compression. Besides weight pruning, we further prune the model at the subword level to exploit the fine-grained subword sparsity for improving the compression performance.
After pruning, the sparse weight matrix of each layer is compressed to a small and dense format by permuting the weights to avoid conflicts and facilitate column packing.
Through weight permutation, the matrix compression rate can be significantly improved, compared to the state-of-the-art compression techniques. As a result, the throughput and energy efficiency can be improved by 2.75 and 1.86 times, respectively.


\bibliographystyle{./IEEEtran}

\begin{thebibliography}{10}
\providecommand{\url}[1]{#1}
\csname url@samestyle\endcsname
\providecommand{\newblock}{\relax}
\providecommand{\bibinfo}[2]{#2}
\providecommand{\BIBentrySTDinterwordspacing}{\spaceskip=0pt\relax}
\providecommand{\BIBentryALTinterwordstretchfactor}{4}
\providecommand{\BIBentryALTinterwordspacing}{\spaceskip=\fontdimen2\font plus
\BIBentryALTinterwordstretchfactor\fontdimen3\font minus
  \fontdimen4\font\relax}
\providecommand{\BIBforeignlanguage}[2]{{%
\expandafter\ifx\csname l@#1\endcsname\relax
\typeout{** WARNING: IEEEtran.bst: No hyphenation pattern has been}%
\typeout{** loaded for the language `#1'. Using the pattern for}%
\typeout{** the default language instead.}%
\else
\language=\csname l@#1\endcsname
\fi
#2}}
\providecommand{\BIBdecl}{\relax}
\BIBdecl

\bibitem{Alexnet}
A.~Krizhevsky, I.~Sutskever, and G.~E. Hinton, ``Imagenet classification with
  deep convolutional neural networks,'' in \emph{Proc. 25th Int. Conf. Neural
  Inf. Process. Syst.}, Dec. 2012, pp. 1097--1105.

\bibitem{Yolo}
J.~Redmon, S.~K. Divvala, R.~B. Girshick, and A.~Farhadi, ``You only look once:
  Unified, real-time object detection,'' \emph{2016 IEEE Conference on Computer
  Vision and Pattern Recognition (CVPR)}, pp. 779--788, 2016.

\bibitem{Unet}
O.~Ronneberger, P.~Fischer, and T.~Brox, ``U-net: Convolutional networks for
  biomedical image segmentation,'' \emph{CoRR}, vol. abs/1505.04597, 2015.

\bibitem{HanCompression}
\BIBentryALTinterwordspacing
S.~Han, J.~Pool, J.~Tran, and W.~J. Dally, ``Learning both weights and
  connections for efficient neural networks,'' \emph{CoRR}, vol.
  abs/1506.02626, 2015. [Online]. Available:
  \url{http://arxiv.org/abs/1506.02626}
\BIBentrySTDinterwordspacing

\bibitem{ToPruneOrNot}
M.~Zhu and S.~Gupta, ``To prune, or not to prune: Exploring the efficacy of
  pruning for model compression,'' \emph{CoRR}, vol. abs/1710.01878, 2017.

\bibitem{TpuGoogle}
\BIBentryALTinterwordspacing
N.~P. Jouppi, C.~Young, N.~Patil \emph{et~al.}, ``In-datacenter performance
  analysis of a tensor processing unit,'' in \emph{Proceedings of the 44th
  Annual International Symposium on Computer Architecture}, ser. ISCA
  '17.\hskip 1em plus 0.5em minus 0.4em\relax New York, NY, USA: ACM, 2017, pp.
  1--12. [Online]. Available: \url{http://doi.acm.org/10.1145/3079856.3080246}
\BIBentrySTDinterwordspacing

\bibitem{PackCNN}
H.~Kung, B.~McDanel, and S.~Q. Zhang, ``Packing sparse convolutional neural
  networks for efficient systolic array implementations: Column combining under
  joint optimization,'' in \emph{Proc. 24th Int. Conf. Archit. Support for
  Program. Lang. \& Oper. Syst. (ASPLOS)}.\hskip 1em plus 0.5em minus
  0.4em\relax New York, NY, USA: ACM, 2019, pp. 821--834.

\bibitem{NetworkSlimming}
Z.~Liu, J.~Li, Z.~Shen, G.~Huang, S.~Yan, and C.~Zhang, ``Learning efficient
  convolutional networks through network slimming,'' \emph{CoRR}, vol.
  abs/1708.06519, 2017.

\bibitem{FilterPruning1}
H.~Li, A.~Kadav, I.~Durdanovic, H.~Samet, and H.~P. Graf, ``Pruning filters for
  efficient convnets,'' \emph{CoRR}, vol. abs/1608.08710, 2016.

\bibitem{FilterPruning2}
\BIBentryALTinterwordspacing
Q.~Huang, S.~K. Zhou, S.~You, and U.~Neumann, ``Learning to prune filters in
  convolutional neural networks,'' \emph{CoRR}, vol. abs/1801.07365, 2018.
  [Online]. Available: \url{http://arxiv.org/abs/1801.07365}
\BIBentrySTDinterwordspacing

\bibitem{TheStateOfSparsity}
T.~Gale, E.~Elsen, and S.~Hooker, ``The state of sparsity in deep neural
  networks,'' \emph{CoRR}, vol. abs/1902.09574, 2019.

\bibitem{TightCompression}
X.~{Chen}, J.~{Zhu}, J.~{Jiang}, and C.~Y. {Tsui}, ``Tight compression:
  Compressing cnn model tightly through unstructured pruning and simulated
  annealing based permutation,'' in \emph{2020 57th ACM/IEEE Design Automation
  Conference (DAC)}, 2020, pp. 1--6.

\bibitem{ImageNet}
O.~Russakovsky, J.~Deng, H.~Su \emph{et~al.}, ``Imagenet large scale visual
  recognition challenge,'' \emph{Int. J. Comput. Vis. (IJCV)}, vol. 115, no.~3,
  pp. 211--252, Dec. 2015.

\bibitem{BiScale}
S.~{Jain}, S.~{Venkataramani}, V.~{Srinivasan}, J.~{Choi}, K.~{Gopalakrishnan},
  and L.~{Chang}, ``Biscaled-dnn: Quantizing long-tailed datastructures with
  two scale factors for deep neural networks,'' in \emph{2019 56th ACM/IEEE
  Design Automation Conference (DAC)}, 2019, pp. 1--6.

\bibitem{OutlierAware}
E.~{Park}, D.~{Kim}, and S.~{Yoo}, ``Energy-efficient neural network
  accelerator based on outlier-aware low-precision computation,'' in \emph{2018
  ACM/IEEE 45th Annual International Symposium on Computer Architecture
  (ISCA)}, 2018, pp. 688--698.

\bibitem{Cifar10}
A.~Krizhevsky, ``Learning multiple layers of features from tiny images,'' Univ.
  of Toronto, Toronto, Canada, Tech. Rep., 2009.

\bibitem{BinPacking}
R.~Rao and S.~Iyengar, ``Bin-packing by simulated annealing,'' \emph{Comput.
  Math. Appl.}, vol.~27, no.~5, pp. 71 -- 82, 1994.

\bibitem{PlacementRoute}
D.~F. Wong, H.~W. Leong, and C.~L. Liu, \emph{Simulated Annealing for VLSI
  Design}.\hskip 1em plus 0.5em minus 0.4em\relax Norwell, MA, USA: Kluwer
  Academic Publishers, 1988.

\bibitem{cacti70}
R.~Balasubramonian, A.~B. Kahng, N.~Muralimanohar, A.~Shafiee, and V.~Srinivas,
  ``Cacti 7: New tools for interconnect exploration in innovative off-chip
  memories,'' \emph{ACM Trans. Archit. Code Optim.}, vol.~14, no.~2, pp.
  14:1--14:25, Jun. 2017.

\bibitem{nangate}
J.~Knudsen, ``Nangate 45nm open cell library,'' in \emph{CDNLive EMEA}, 2008.

\bibitem{ShiftNet}
B.~Wu, A.~Wan, X.~Yue \emph{et~al.}, ``Shift: {A} zero flop, zero parameter
  alternative to spatial convolutions,'' \emph{CoRR}, vol. abs/1711.08141,
  2017.

\end{thebibliography}



\begin{IEEEbiography}[{\includegraphics[width=1in,height=1.25in,clip,keepaspectratio]{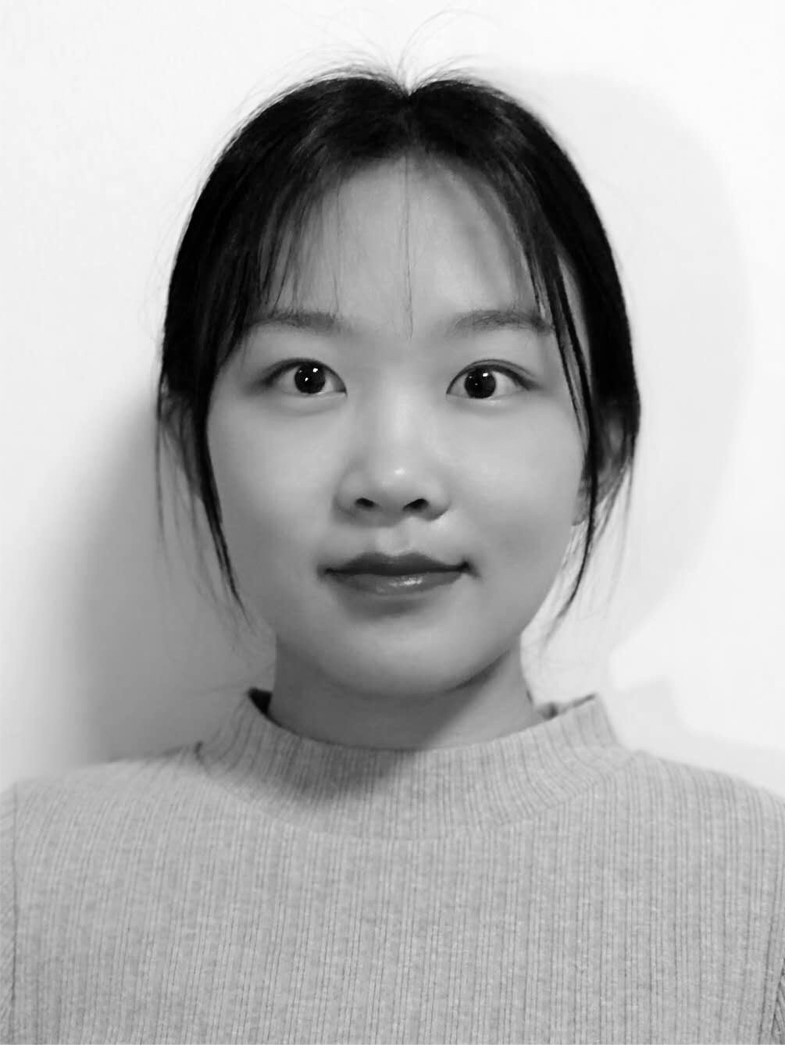}}]{Xizi Chen}
received her B.E. degree in Electronic Science and Technology 
from Xi'an Jiao Tong University, Xi'an, China, in 2015. 
She is currently working towards her Ph.D. degree in Electronic and 
Computer Engineering at Hong Kong University of Science and Technology (HKUST), Hong Kong, China. 

Her current research interests are in hardware-software codesign for high-throughput and energy-efficient deep learning accelerators.
\end{IEEEbiography}

\begin{IEEEbiography}[{\includegraphics[width=1in,height=1.25in,clip,keepaspectratio]{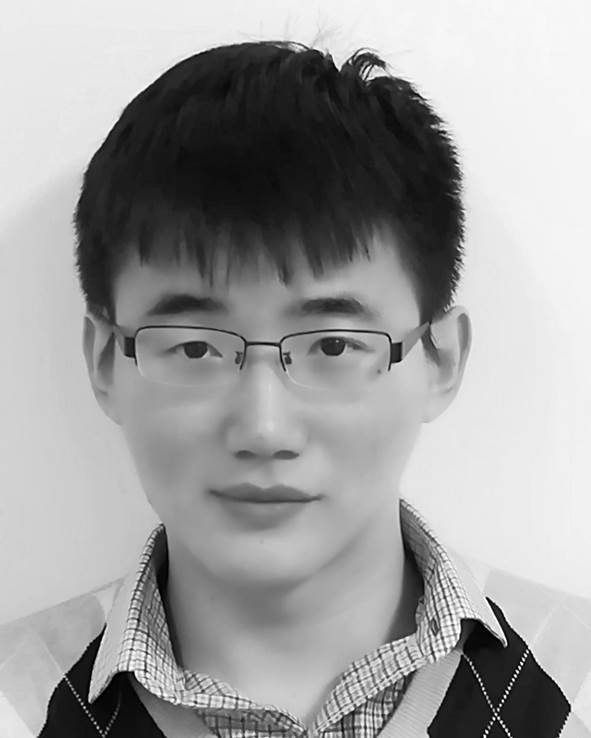}}]{Jingyang Zhu}
received his B.S. degree in School of Microelectronics from the Shanghai Jiao Tong University, 
Shanghai, China, in 2013, and his Ph.D. degree in Electronic and Computer Engineering from the 
Hong Kong University of Science and Technology, Hong Kong, in 2019. He is currently an architect 
with NVIDIA, Shanghai. 

His research interests are in computer architecture, low-power VLSI implementation and 
domain specific architecture for deep learning.
\end{IEEEbiography}

\begin{IEEEbiography}[{\includegraphics[width=1in,height=1.25in,clip,keepaspectratio]{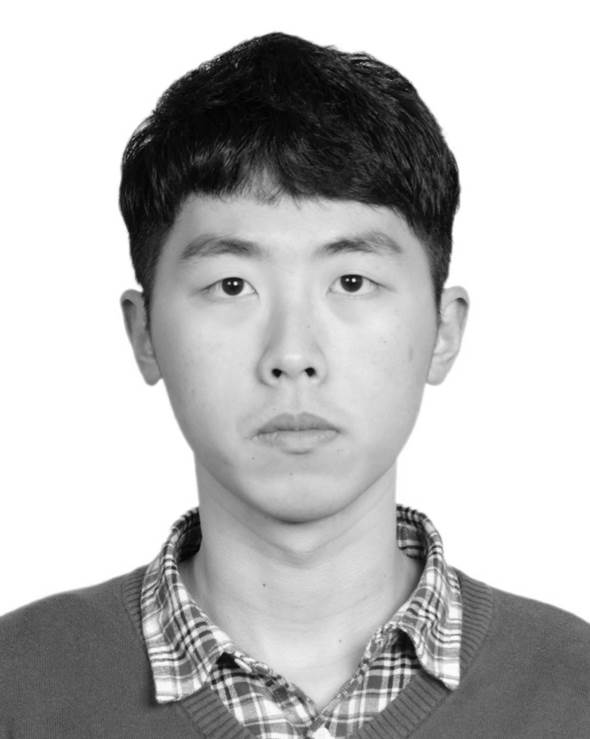}}]{Jingbo Jiang}
received his B.S. degree from Hefei University of Technology, Hefei, China, in 2016, and
his M.S. degree from the Hong Kong University of Science and Technology, Hong Kong, in 2020.
He is currently working towards his Ph.D. degree at Hong Kong University of 
Science and Technology, Hong Kong, China.

His research interests are in circuit design for implementing deep learning algorithms.
\end{IEEEbiography}

\vfill

\begin{IEEEbiography}[{\includegraphics[width=1in,height=1.25in,clip,keepaspectratio]{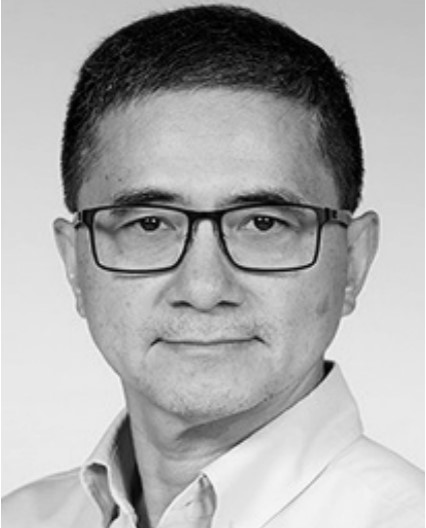}}]{Chi-Ying Tsui}
(Senior Member, IEEE) received his B.S. degree in electrical engineering from
the University of Hong Kong, Hong Kong, in 1982, and his Ph.D. degree in computer engineering
from the University of Southern California, Los Angeles, CA, USA, in 1994.

In 1994, he joined the Department of Electronic and Computer Engineering, 
The Hong Kong University of Science and Technology, Hong Kong, where he is 
currently a Full Professor. He has authored more than 170 referred publications. 
He holds ten U.S. patents on power management, VLSI, and multimedia systems.
His research interests include designing VLSI architectures for low-power
multimedia and wireless applications, developing power management circuits
and techniques for embedded portable devices and ultralow power systems.

Dr. Tsui received the Best Paper Awards from the \textsc{IEEE Transactions
on Very Large Scale Integration (VLSI) Systems} in 1995, the IEEE ISCAS in 1999, the IEEE/ACM ISLPED in 2007,
the IEEE DELTA in 2008, and CODES in 2012. He also received the Design
Awards at the IEEE ASP-DAC University Design Contest in 2004 and 2006. 
\end{IEEEbiography}
\vfill

\end{document}